\documentclass[a4paper,fleqn]{cas-dc}

\usepackage[numbers]{natbib}
\usepackage{hyperref}
\usepackage{xcolor}
\hypersetup{colorlinks=true, citecolor=blue, linkcolor=blue, urlcolor=blue}

\def\tsc#1{\csdef{#1}{\textsc{\lowercase{#1}}\xspace}}
\tsc{WGM}
\tsc{QE}

\begin{document}
\let\WriteBookmarks\relax
\def\floatpagepagefraction{1}
\def\textpagefraction{.001}
\let\printorcid\relax 

\shorttitle{Weakly Supervised Camouflaged Object Detection Based on the SAM Model and Mask Guidance}    

\shortauthors{Xia Li et al.}

\title[mode = title]{Weakly Supervised Camouflaged Object Detection Based on the SAM Model and Mask Guidance}

\author[1]{Xia Li}
\fnmark[1] 
\ead{lx7211@stu.ouc.edu.cn} 

\author[1]{Xinran Liu}
\fnmark[2] 
\ead{lxr7766@stu.ouc.edu.cn}

\author[1]{Lin Qi}
\fnmark[3] 
\cormark[1]
\ead{qilin@ouc.edu.cn}

\author[1]{Junyu Dong}
\fnmark[4] 
\ead{dongjunyu@ouc.edu.cn}

\address[1]{School of Computer Science and Technology, Ocean University of China, Qingdao 266100, China}

\cortext[1]{Corresponding author} 
\begin{abstract}
Camouflaged object detection (COD) from a single image is a challenging task due to the high similarity between objects and their surroundings. Existing fully supervised methods require labor-intensive pixel-level annotations, making weakly supervised methods a viable compromise that balances accuracy and annotation efficiency. However, weakly supervised methods often experience performance degradation due to the use of coarse annotations. In this paper, we introduce a new weakly supervised approach for camouflaged object detection to overcome these limitations. Specifically, we propose a novel network, MGNet, which tackles edge ambiguity and missed detections by utilizing initial masks generated by our custom-designed Cascaded Mask Decoder (CMD) to guide the segmentation process and enhance edge predictions. We introduce a Context Enhancement Module (CEM) to reduce the missing detection, and a Mask-guided Feature Aggregation Module (MFAM) for effective feature aggregation. For the weak supervision challenge, we propose BoxSAM, which leverages the Segment Anything Model (SAM) with bounding-box prompts to generate pseudo-labels. By employing a redundant processing strategy, high quality pixel-level pseudo-labels are provided for training MGNet. Extensive experiments demonstrate that our method delivers competitive performance against current state-of-the-art methods.
\end{abstract}

\begin{keywords}
Camouflaged object detection \sep 
Weak supervision \sep 
Segment anything model \sep
Mask guidance
\end{keywords}
\maketitle
\section{Introduction}
Camouflaged object detection (COD) refers to the pixel-level segmentation of objects that blend seamlessly with their surrounding environment \cite{fan2020camouflaged}. This task presents greater challenges compared to general object detection and salient object detection (SOD) because camouflaged objects usually make themselves highly match their surroundings in terms of texture, color and shape. The detection of camouflaged objects finds extensive utility in various practical domains such as polyp segmentation \cite{fan2020pranet} in the medical domain, pest detection \cite{fuentes2017robust} in agriculture and object surface defect detection in the industrial field \cite{fan2023advances}. 

\begin{figure} 
\centerline{\includegraphics[width=\columnwidth]{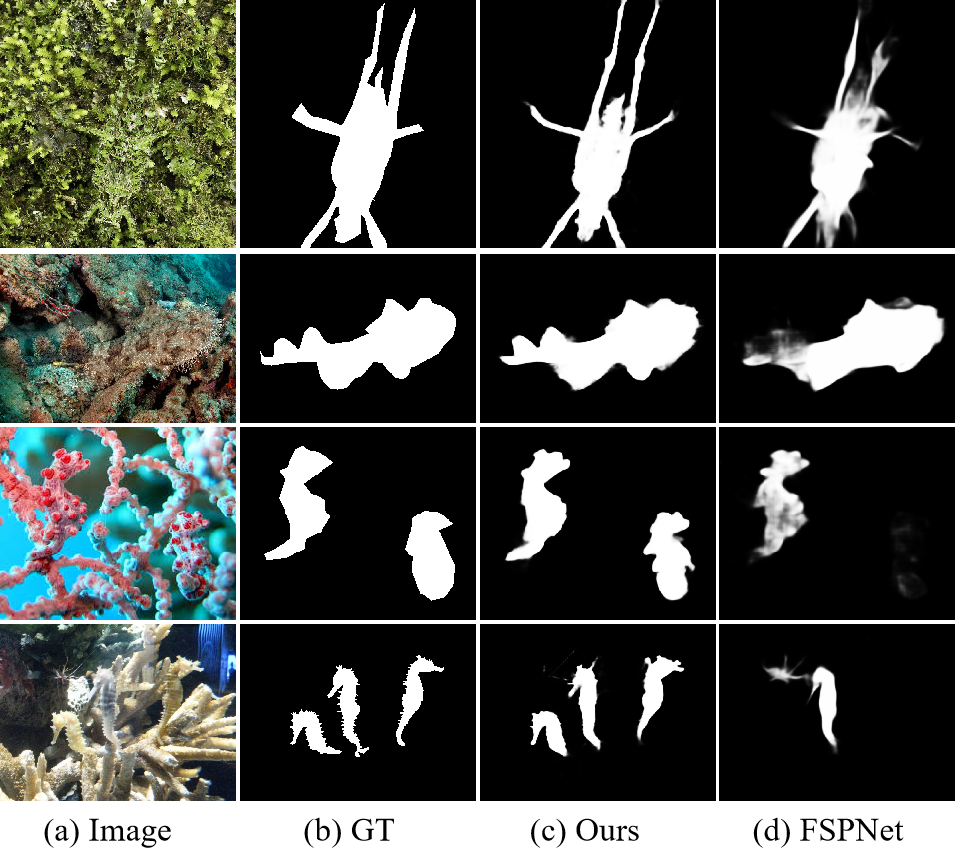}}
\caption{Visual examples of different methods. (a) RGB images. (b) Ground truth. (c)-(d) The detection results obtained by (c) Our results, (d) FSPNet \cite{huang2023feature}.}
\label{fig:fig1} 
\end{figure}

Benefiting from the strong learning capacity of advanced CNN and transformer architectures, COD has achieved remarkable progress in recent years. The task of COD has evolved from relying on low-level visual features, such as traditional texture \cite{song2010new}, gradient \cite{pan2011study}, and motion features \cite{hou2011detection}, to extracting high-level semantic features using deep learning methods. Recently, numerous deep learning-based methods for COD have achieved significant advancements \cite{huang2023feature, hu2023high, liu2023mscaf, he2023strategic}. However, our experiments reveal that these methods still exhibit two shortcomings: 1)  the ambiguous edges of camouflaged objects. 2) the missed detection of camouflaged objects. As shown in Figure \ref{fig:fig1}, due to the intrinsic similarity between the camouflaged objects and the backgrounds, FSPNet \cite{huang2023feature} fails to completely detect the edges of caddis moth (row 1) and fish (row 2), and even misses the small hippocampi (rows 3 and 4). 

In addition to these two problems, fully supervised COD training typically relies on pixel-level supervision, which involves costly pixel-level annotation \cite{bearman2016s}. Consequently, weakly supervised COD holds significant research importance. Weakly supervised camouflaged object detection (WSCOD) relies on point annotations, bounding-box annotations or scribble annotations in the foreground or background, eliminating the need for pixel-level labeling. This approach enhances labeling efficiency and reduces costs. However, coarse annotations that only differentiate between foreground and background result in degraded segmentation performance. For instance, WSCOD based on scribble annotations \cite{he2023weakly} results in decreased segmentation accuracy. In addition, the community of weakly supervised salient object detection \cite{zhang2020weakly, yu2021structure, liang2022tree, liu2021weakly, wang2023weakly, Liu2024weakly} has made remarkable advancements, whereas there have been only very few attempts in WSCOD \cite{he2023weakly, he2024weakly, chen2024sam}. How to balance accuracy and annotation efficiency in WSCOD is still under exploration.

Inspired by the process of human observation, we first roughly localize the position and contour of the camouflaged object and then refine its edges. We propose the Mask-guided Network (MGNet) for the COD task, which guides subsequent segmentation through the initial generated masks by our designed Cascaded Mask Decoder (CMD). Specifically, to reduce the missing detection, we design a Context Enhancement Module (CEM), which employs convolutional layers with varying dilation rates to reduce information loss during the downsampling process. Besides, the CMD incrementally fuses multi-scale features, including high-level semantic information and low-level detail information, to guide subsequent segmentation from a global perspective. To address the ambiguous edges problem, we design a Mask-guided Feature Aggregation Module (MFAM), which is utilized to identify the fine regions of objects by fusing adjacent-level features with the guidance of masks from the CMD.

To address the challenge of weak supervision, we propose BoxSAM. We annotate the images with bounding boxes, which are less affected by human subjective factors compared to scribble annotations and point annotations. This is attributable to the fact that both the position of the point annotations and the length and location of the scribble annotations can significantly impact the performance of the model. In contrast, under ideal circumstances, a bounding box annotation represents the minimal enclosing box that completely encompasses the object. Segment Anything Model (SAM) \cite{kirillov2023segment} possesses powerful segmentation capabilities, but its performance in the field of camouflaged object detection is not particularly outstanding due to the high intrinsic similarity between the camouflaged objects and the backgrounds, segmentation results of SAM may mistakenly classify the backgrounds as camouflaged objects \cite{ji2304sam, tang2023can, ji2024segment}. 
To address this issue, we design a redundancy processing strategy to refine the initial labels, providing high-quality pixel-level pseudo labels for training MGNet. In contrast to the strategies during the training period \cite{liu2022adaptive}, we directly process the pseudo-labels generated by SAM based on the bounding-box prompts. We propose the redundancy processing strategy specifically designed to address the issue of redundant predictions generated by SAM when segmenting images with cluttered backgrounds or scenarios where the background and camouflaged objects exhibit high similarity. By integrating deep learning model MGNet with the generalization capabilities of SAM, we achieve more accurate pseudo-label generation. 

Our contributions in this paper are as follows:
\begin{itemize}
\item The proposed BoxSAM is a WSCOD method that generates pseudo-labels by annotating bounding-boxes of the camouflaged objects and integrating the SAM model. Furthermore, we propose a redundancy processing strategy to obtain more accurate pseudo-labels.

\item We propose a novel Mask-guided Network (MGNet) that enhances subsequent segmentation by utilizing the initially generated masks as guidance. We propose a CEM to reduce the missing detection. Additionally, we design a MFAM to aggregate features at different levels under the guidance of masks generated by the CMD.

\item BoxSAM is evaluated on three widely used COD datasets, the experiments demonstrate that our method achieves state-of-the-art performance. MGNet outperforms the current state-of-the-art COD methods. In addition, it also demonstrates excellent performance in COD-related applications.
\end{itemize}

\section{Related work}
\subsection{Camouflaged object detection}
Traditional COD methods are susceptible to noise, resulting in relatively poor detection performance. With the advancement of deep learning, many contemporary COD methods now leverage deep learning techniques. Fan et al. \cite{fan2020camouflaged} simulated the hunting process and proposed SINet, which initially searched for the disguised objects through the receptive field module, and then detected them through part of the decoder component and the search attention module. Similarly, Mei et al. \cite{mei2023distraction} imitated the natural predation process to locate potential objects in a global perspective, and then used the focusing module for recognition to gradually refine the initial predictions by focusing on the blurred regions. Liu et al. \cite{liu2024search} designed a region search module to mimic predator behavior to locate potential object regions, enhancing object location detection. Ye et al. \cite{ye2024reverse} designed a diverse feature enhancement module that simulates the correspondingly expanded receptive fields of the human visual system by using convolutional kernels with different dilation rates in parallel.  Pang et al. \cite{pang2024zoomnext} proposed an effective unified collaborative pyramid network, which emulates the way humans behave when they view unclear images and videos, specifically the action of magnifying and reducing the view.

Some methods introduce additional priors. Xu et al. \cite{xu2021boundary} explored semantic clues based on edge truth values and combined edge features with semantic features through an edge-guided fusion module. However, in a complex environment, the boundary leads to noise. To address this, Ji et al. \cite{ji2023deep} utilized a gradient-guided transfer module to aggregate multi-source features. Liu et al. \cite{liu2023mscaf} designed a Dense Interactive Decoder module that produces a rough localization map to enhance subsequent fusion features for more accurate detection. Yue et al. \cite{yue2023dual} focused on object regions and edges for detecting camouflaged objects in a coarse-to-fine manner. 

Zhao et al. \cite{zhao2024focusdiffuser} proposed a denoising diffusion model to investigate how generative models can enhance the detection and interpretation of camouflaged objects. DiffCOD \cite{chen2023diffusion} is also based on the diffusion model, which regards the task of COD as a denoising diffusion process which progresses from noisy masks to object masks. In addition, many scholars have also explored multi-image collaborative methods. Yu et al. \cite{yu2024exploring} proposed  DSAM by combining depth maps. They utilized the zero-shot capability of SAM to realize precise segmentation in the RGB-D domain. From the perspective of multi-scale context aggregation, He et al. \cite{he2024weakly} designed a multi-scale feature grouping module that first groups features at different granularities and then aggregates these grouping results. For high-resolution dichotomous image segmentation, Zheng et al. \cite{zheng2024bilateral} proposed a novel bilateral reference framework (BiRefNet), which achieved excellent segmentation results in the COD task.

Existing methods for COD have demonstrated good performance. However, issues such as the missing detection and edge ambiguity persist, as illustrated in Figure \ref{fig:fig1}. To address these issues, we design a Mask-guided Network (MGNet) that guides subsequent segmentation through the initial generated masks.

\begin{figure*}[b]
\centerline{\includegraphics[width=1\textwidth]{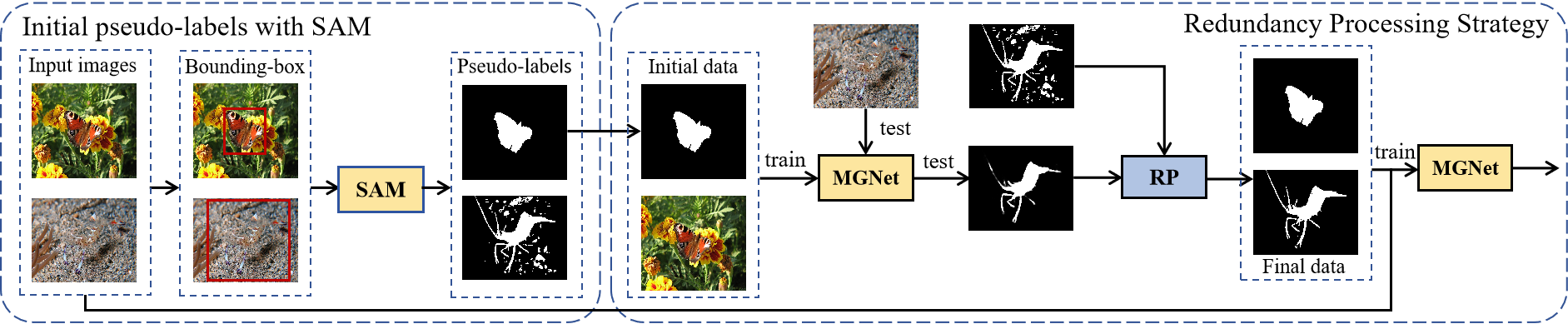}}
\caption{Overview of WSCOD with bounding-box supervision (BoxSAM). By annotating the bounding-boxes of the camouflaged objects and combining the SAM \cite{kirillov2023segment} model to output the pseudo-labels, we design a redundancy processing strategy (see Section \ref{RPS}) with MGNet (see Section \ref{DD}) to process the pseudo-labels. }
\label{fig:SAM}
\end{figure*}
\begin{figure}
\centerline{\includegraphics[width=\columnwidth]{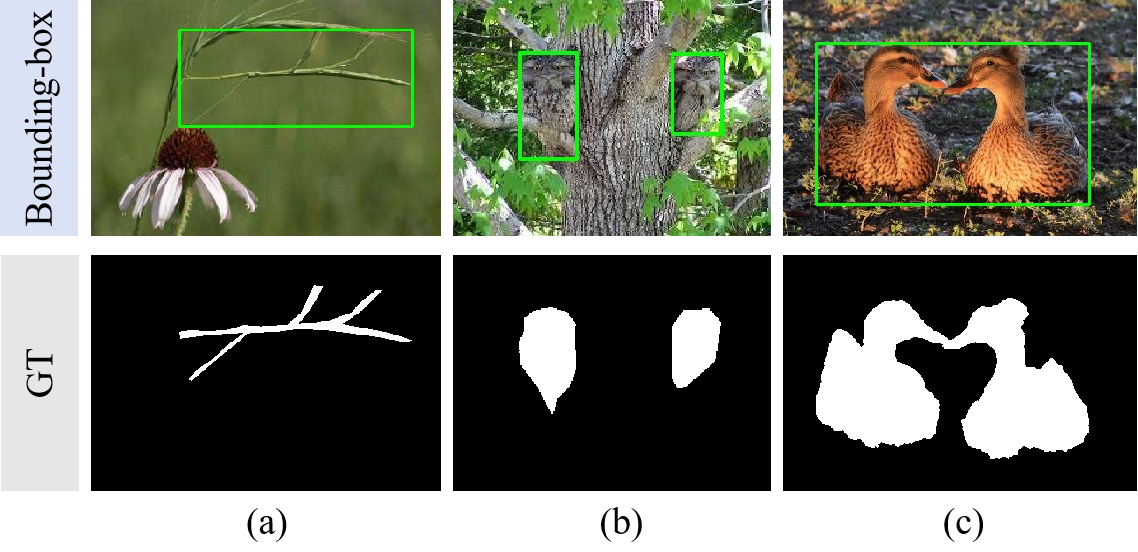}}
\caption{Bounding-box annotations method, images from CAMO \cite{le2019anabranch} and COD10K  \cite{fan2020camouflaged}. The green bounding-boxes represent our annotated bounding-boxes. The annotations are classified into 3 cases: (a) A  bounding-box contains a camouflaged object. (b) Multiple bounding-boxes contain multiple camouflaged objects. (c) A bounding-box contains multiple camouflaged objects.}
\label{fig:bounding_box}
\end{figure}

\subsection{Weakly supervised COD}
To balance labeling costs with detection performance, many researchers have turned to WSCOD. He et al. \cite{he2023weakly} proposed the first WSCOD framework, though its segmentation performance is limited by the sparse information provided by scribble annotations. With the advent of large models, He et al. \cite{he2024weakly} proposed a WSCOD method leveraging a large model. This approach inputs images with scribble annotations into Segment Anything Model (SAM) \cite{kirillov2023segment}, employing multiple enhancement result integration, entropy-based pixel-level weighting and entropy-based image-level selection strategies to provide more accurate supervision for the segmentation model. Chen et al. \cite{chen2024sam} proposed SAM-COD, which is designed to accommodate arbitrary weakly supervised labels. Additionally, in terms of learning from noisy pseudo-labels, Liu et al. \cite{liu2022adaptive} proposed ADELE for semantic segmentation in the presence of noisy pixel-level annotations, which detects the beginning of the memorization phase separately for each category to adaptively correct noisy annotations during training and incorporates a regularization term.

Methods that rely on scribble supervision are influenced by human subjective factors, such as the position of the marker and the length of the scribble line, which can negatively impact the segmentation results. Compared to scribble annotations, the bounding-box annotations are more intuitive and concise, and are less influenced by subjective factors. Therefore, we implement WSCOD using bounding-box annotations combined with SAM.

\subsection{Weakly supervised SOD}
SOD aims to segment regions that attract human attention \cite{wang2021salient}, typically characterized by clearer edges compared to camouflaged objects. Research on weakly supervised SOD began earlier compared to WSCOD. Zhang et al. \cite{zhang2020weakly} utilized scribble annotations to learn saliency and designed a scribble enhancement scheme to iteratively refine these annotations, which were subsequently used to supervise the learning of high-quality saliency maps. To streamline training complexity, Yu et al. \cite{yu2021structure} devised SCWS, which is an end-to-end training approach for predicting comprehensive salient regions with complete object structures. This method propagates labels to unlabeled regions using image features and pixel distances, eliminating the need for preprocessing, post-processing, and extra supervised data. Addressing coarse annotated semantic segmentation, Liang et al. \cite{liang2022tree} introduced a tree energy loss mechanism that offers semantic guidance to unlabeled pixels.

In addition to scribble annotations, bounding-box annotations are utilized in weakly supervised SOD. Liu et al. \cite{liu2021weakly} predicted the pixel-level pseudo ground truth saliency map using saliency bounding-boxes, generated initial saliency maps via an unsupervised SOD method, and iteratively refined the initial pseudo ground truth through a multi-task graph refinement network. Wang et al. \cite{wang2023weakly} integrated bounding-box labels with the GrabCut algorithm to produce the initial saliency map and devise a correction module to enhance it. Liu et al. \cite{Liu2024weakly} employed bounding-box and SAM to generate the initial pseudo-labels, applying various strategies to enhance the segmentation accuracy of this map. Liu et al. \cite{liu2025ssfam} designed modal-aware modulators to extract modality-specific knowledge and a siamese decoder to address the discrepancy between training with scribble prompts and testing without prompts.

Due to the strong segmentation capabilities of SAM, it is frequently employed in image segmentation tasks \cite{zhang2023input, chen2024ma}. However, SAM faces significant challenges in camouflaged object detection task. Extensive experiments have demonstrated that SAM encounters challenges in accurately segmenting camouflaged objects \cite{ji2304sam}, often misidentifying backgrounds as camouflaged objects, thereby introducing redundant information. To address this issue, we propose a redundancy processing strategy to enhance the performance of WSCOD.

\section{Methodology}
Figure \ref{fig:SAM} shows the pipeline of our proposed method, we propose a WSCOD method that utilizes bounding-box annotations as prompts in combination with SAM. In order to solve the problem of redundant information in the preliminary segmentation results, we develop a Redundancy Processing Strategy (RPS) that incorporates our designed Mask-guided Network (MGNet) which includes three proposed modules: the Cascaded Mask Decoder (CMD), the Context Enhancement Module (CEM) and the Mask-guided Feature Aggregation Module (MFAM). Detailed explanations of each component are provided in Sections \ref{box}--\ref{DD}.

\begin{figure*}[b]
\centerline{\includegraphics[width=1\textwidth]{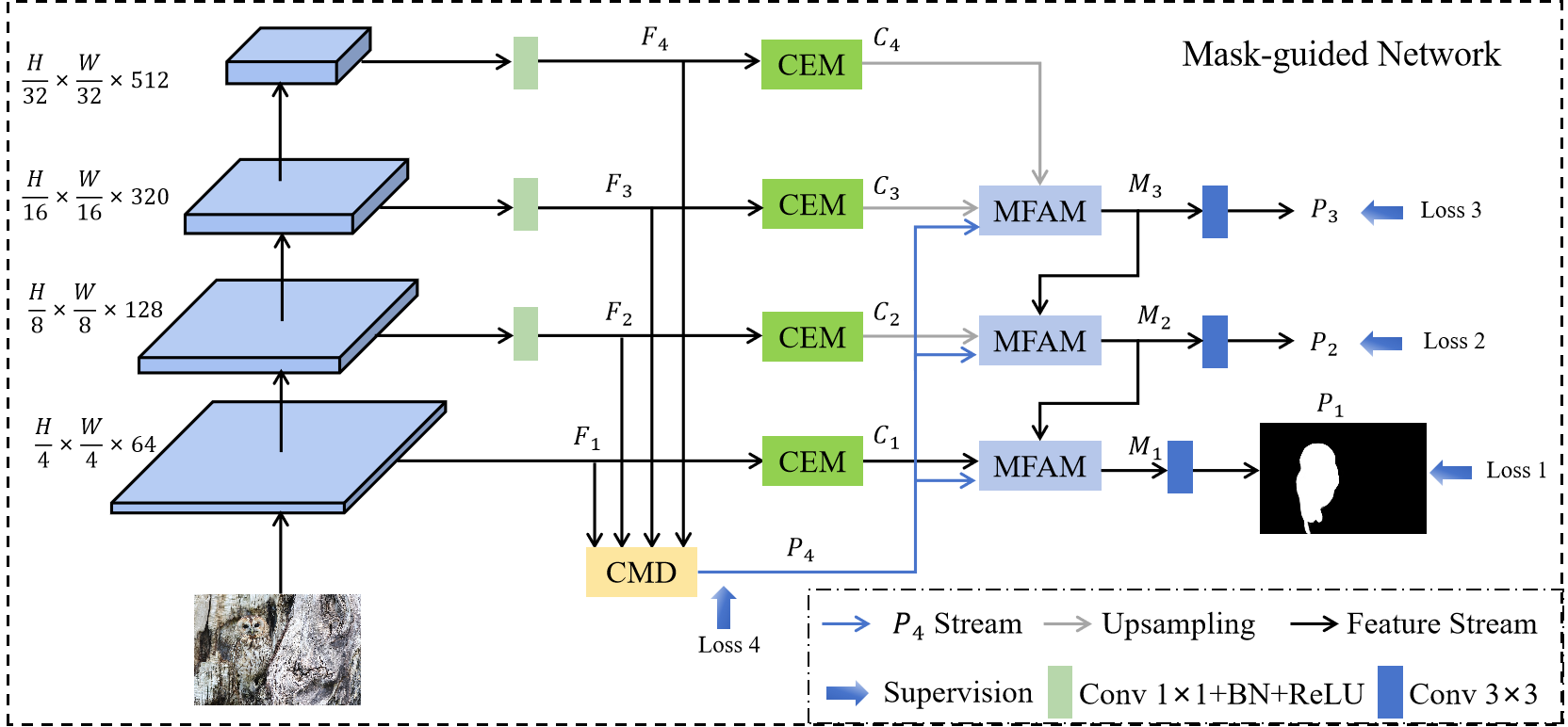}}
\caption{Overview of Mask-guided Network (MGNet). The MGNet consists of our designed Cascaded Mask Decoder (CMD, see Section \ref{AA}), Context Enhancement Module (CEM, see Section \ref{BB}) and Mask-guided Feature Aggregation Module (MFAM, see Section \ref{CC}).}
\label{fig:MGNetwork}
\end{figure*}
\begin{figure}  
\centerline{\includegraphics[width=\columnwidth]{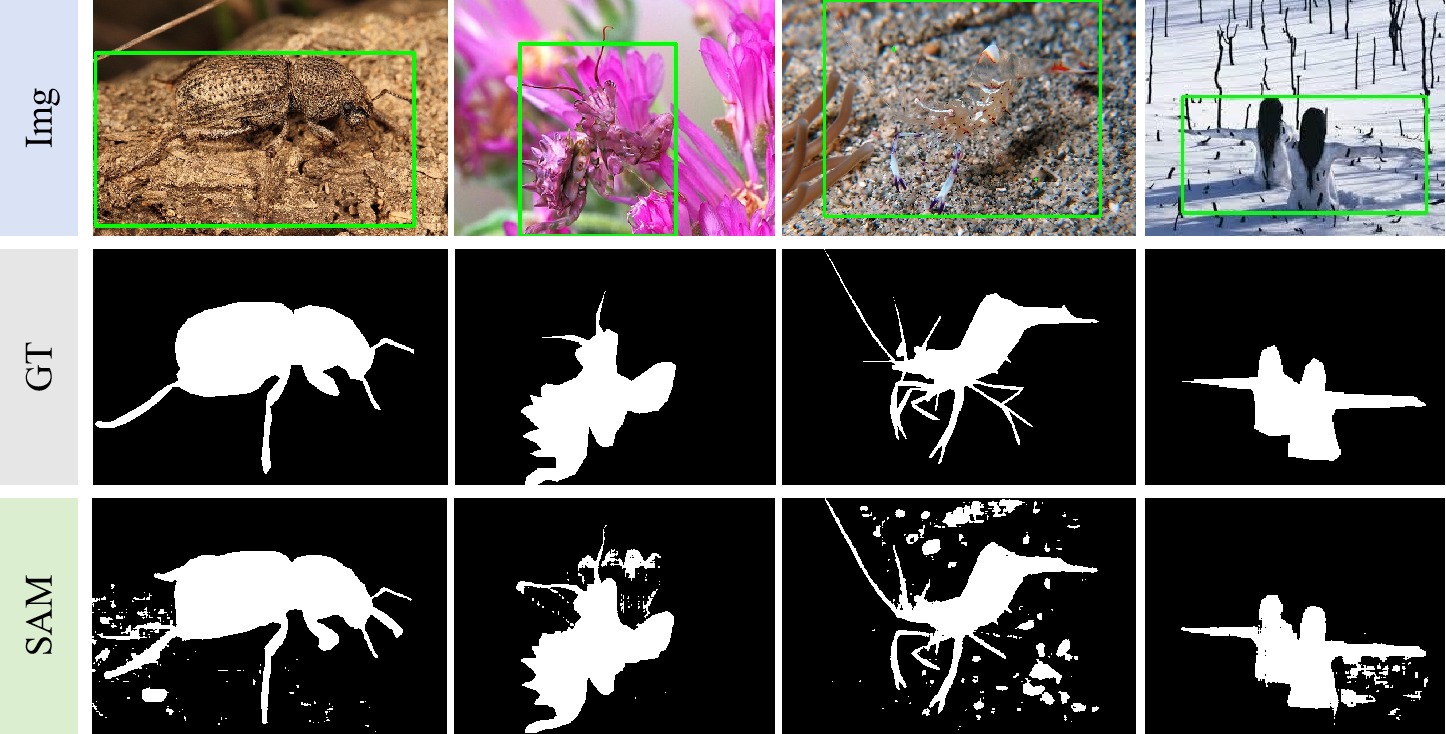}}
\caption{Some examples of SAM segmentation with bounding-boxes.}
\label{fig:F}
\end{figure}

\subsection{Initial pseudo-labels with SAM}\label{box}
The Segment Anything Model (SAM) \cite{kirillov2023segment} is a newly introduced model for visual segmentation, trained with more than one billion segmentation masks. It has demonstrated exceptional performance in generating accurate segmentation masks across a broad range of object categories. SAM draws on the prompt strategy in the Natural Language Processing (NLP) field to complete the segmentation of arbitrary objects by providing prompts to the image segmentation task. It can be prompted by point annotations, scribble annotations and bounding-box annotations. In this paper, we use bounding-box annotations as prompts.

As shown in Figure \ref{fig:bounding_box}, we annotate the bounding-boxes of the camouflaged objects according to the following rules \cite{Liu2024weakly}: 1) When an image only has a camouflaged object, the smallest bounding-box that encompasses the object is annotated, as shown in (a). 2) When an image has multiple camouflaged objects, each object is annotated separately according to rule 1), as shown in (b). If there is overlap between multiple camouflaged objects, we merge them into a bounding-box, as shown in (c).

As shown in Figure \ref{fig:SAM}, we first annotate the bounding-boxes of the camouflaged objects based on the above rules. Given a camouflaged image $I$ with bounding-box $B$, we generate the initial pseudo-labels $M$ with the bounding-boxes as prompts for the SAM model. 
This process can be formulated as Eq.\ref{eq:SAM}:
\begin{align}
\label{eq:SAM}
&M=SAM(I, B)
\end{align}
where SAM denotes the Segment Anything Model \cite{kirillov2023segment}.

\subsection{Redundancy processing strategy}\label{RPS}
Due to the high similarity between the camouflaged objects and their backgrounds, SAM often mistakes the backgrounds for the camouflaged objects during segmentation \cite{ji2304sam}, as shown in Figure \ref{fig:F}. The quality of the labels significantly affects the model's performance, making it crucial to select accurate pseudo-labels. For each initial pseudo-label, the number of bounding-boxes is calculated following the rule of \cite{Liu2024weakly}. The pseudo-labels whose number of bounding-boxes is equal to the original annotation are used as the initial training data to train the MGNet, while those with mismatched numbers are denoted as $F$. The images corresponding to $F$ are used as test data input to the already trained MGNet. To eliminate redundant information in the initial labels, we use the predicted masks to approximate the position and outline of the camouflaged objects. We save the predicted masks in image format according to \cite{hu2023high,fan2021concealed}, denoted as the segmentation result $P$. To mitigate the redundant information present in $F$, we proposed a redundancy processing (RP) method. The following section provides a  description of the RP method. The connected regions $F_{i}$ of each $F$ are counted. For each $i$, $(m,n)$ represents the pixel coordinates in $F_{i}$. The corresponding region of $F_{i}$ in $P$ is determined by traversing $P(m,n)$. If the set of values of $P(m,n)$ contains non-zero values, the pixel values of $F(m,n)$ are set to 255. If all values in  are 0, the pixel values of $F(m,n)$ are set to 0.  This process can be formulated as Eq.\ref{RPE}:
\begin{align}
\label{RPE}
F(m,n)=\left\{
\begin{aligned}
255 & , & \exists (m,n) \in F_{i}, P(m,n)>0,\\
0 & , & \forall (m,n) \in F_{i}, P(m,n)=0.
\end{aligned}
\right.
\end{align}
where $(m,n)$ denotes the pixel coordinate. Finally, both the initial training dataset and the processed dataset are used to retrain MGNet. The whole process is shown in Figure \ref{fig:SAM}.

\begin{figure}  
\centerline{\includegraphics[width=\columnwidth]{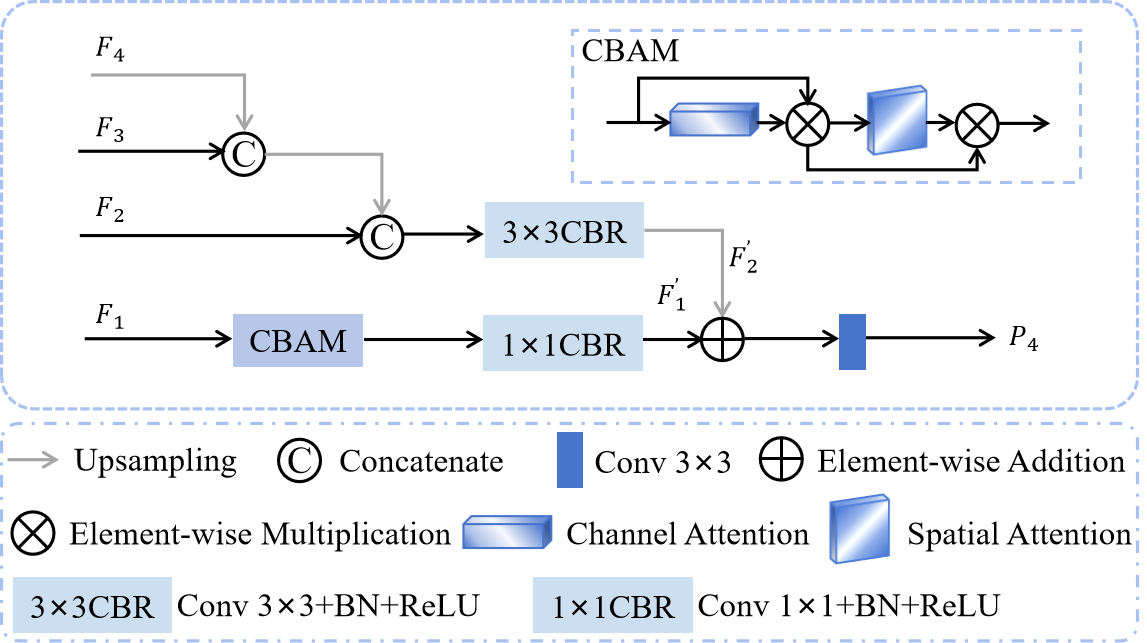}}
\caption{The details of Cascaded Mask Decoder (CMD).}
\label{fig:CMD}
\end{figure}

\subsection{Mask-guided network}\label{DD}
\subsubsection{Overall architecture}
Figure \ref{fig:MGNetwork} shows the structure of our proposed network. We select the Pyramid Visual Transformer (PVT) \cite{wang2021pyramid} as our feature extraction module. Specifically, for a given input image $I \in \mathbb{R}^{H \times W \times 3}$, PVTv2 \cite{wang2022pvt} is used to encode the input image, obtaining multi-level features $F_{i}$ ($i \in \{1, 2, 3, 4\}$) that contain rich spatial details and semantic information. The multi-level features are processed through the Cascaded Mask Decoder (CMD), which progressively integrates information from each layer to generate the mask $P_{4}$. The multi-level features introduce rich context information through the Context Enhancement Module (CEM) to enhance feature representation, resulting in $C_{i}$ ($i \in \{1, 2, 3, 4\}$). Subsequently, under the guidance of mask $P_{4}$, the Mask-guided Feature Aggregation Module (MFAM) aggregates different levels of features and outputs $M_{i}$ ($i \in \{1, 2, 3\}$), where $M_1$ is used for the final output. Detailed explanations of each component are provided in Sections \ref{AA}--\ref{CC}.

\subsubsection{Cascaded mask decoder}\label{AA}
In the task of COD, multi-scale information is believed to be important in distinguishing the targets from their surroundings \cite{xu2023guided}. To fully utilize information at different scales, we design the Cascaded Mask Decoder (CMD), as shown in Figure \ref{fig:CMD}. This module gradually fuses multi-scale features through multi-level processing, integrating high-level semantic information and low-level detail information to generate an initial mask. The initial mask is subsequently used as input in the following steps to progressively refine the details of the camouflaged objects.

High-level semantic information is obtained through upsampling and channel concatenation, resulting in the high-level feature  $F'_{2}$. To capture the details of the camouflaged objects from different dimensions, low-level information $F_{1}$ is processed using the Convolutional Block Attention Module (CBAM) \cite{woo2018cbam} to obtain $F'_{1}$. Finally, the mask $P'_{4}$ is obtained by summing $F'_{1}$ and $F'_{2}$. This process can be formulated as Eq.\ref{eq:cascaded1}, Eq.\ref{eq:cascaded2} and Eq.\ref{eq:cascaded3}:
\begin{align}
\label{eq:cascaded1}    
&F'_{1}=Conv_{1}(CBAM(F_{1}))\\
\label{eq:cascaded2}
&F'_{2}=Conv_{3}(cat((cat(F_{4} \uparrow ,F_{3}))\uparrow,F_{2}))\\
\label{eq:cascaded3}
&P'_{4}=F'_{1}+F'_{2} \uparrow
\end{align}
where $CBAM(\cdot)$ denotes Convolutional Block Attention Module, $\uparrow$ denotes the upsampling, $Conv_{1}(\cdot)$ denotes a $1\times1$ convolution with batch normalization and ReLU. $Conv_{3}(\cdot)$ denotes a $3\times3$ convolution with batch normalization and ReLU. $cat(\cdot)$ represents the concatenation operation. The prediction mask $P_{4}$ can be obtained by processing $P'_{4}$ with a $3\times3$ convolution operation.

\begin{figure}  
\centerline{\includegraphics[width=\columnwidth]{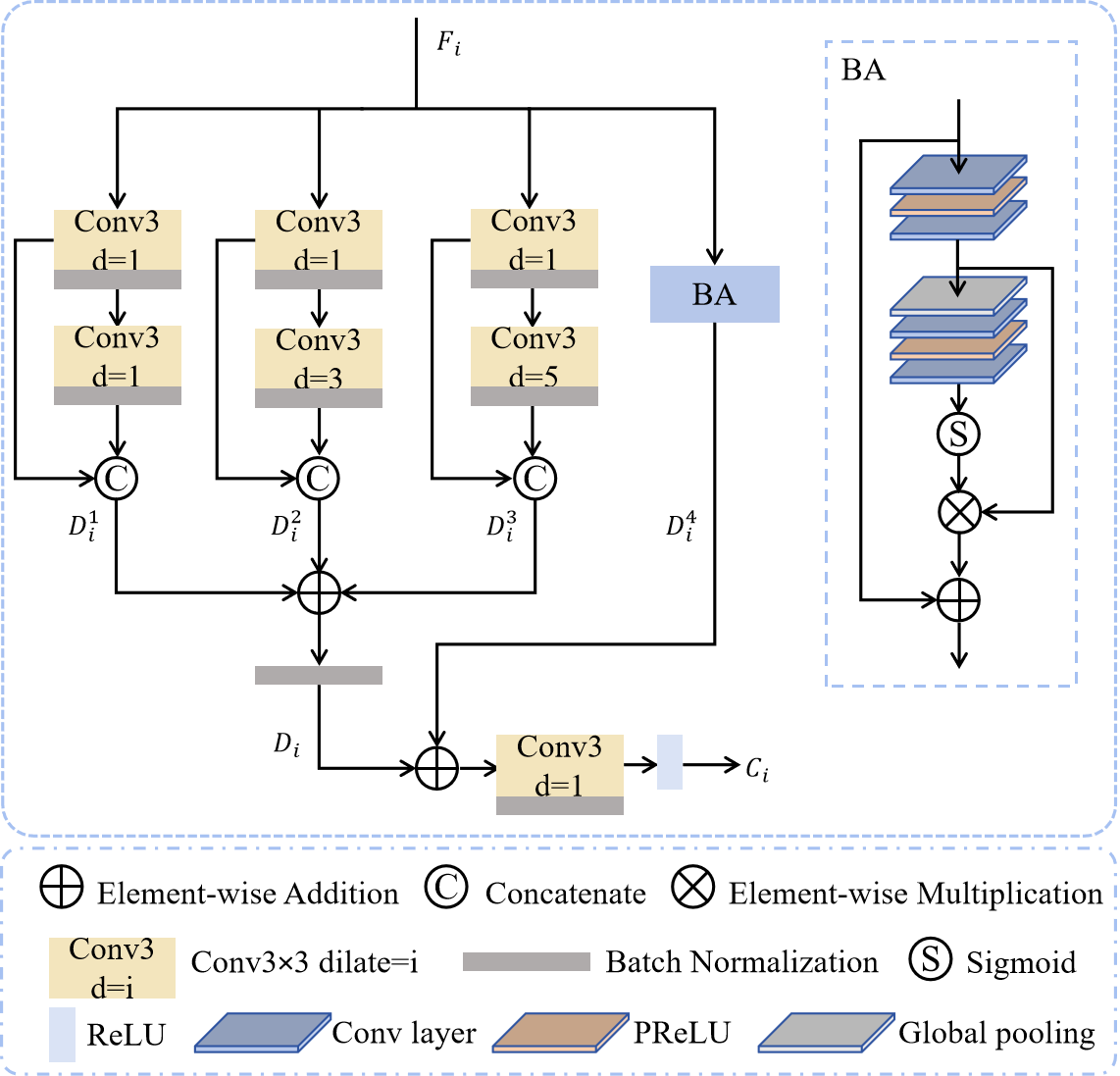}}
\caption{The details of Context Enhancement Module (CEM).}
\label{fig:CEM}
\end{figure}

\subsubsection{Context enhancement module}\label{BB}
In the encoder stage, as the feature space is downsampled, this process poses the risk of losing details or missing camouflaged objects. To this end, we design the Context Enhancement Module (CEM), which employs convolutional layers with varying dilation rates \cite{li2018csrnet} to expand the receptive field and introduce rich context information, thereby compensating for information loss during the downsampling process. Additionally, to further refine the details of camouflaged objects, we introduce the BA \cite{hu2023high}, which replaces ReLU with PReLU in Residual Channel Attention Block (RCAB) \cite{zhang2018image}. The BA focuses on extracting high-frequency information in the image and strengthens attention to details.

As illustrated in Figure \ref{fig:CEM}, the CEM primarily consists of four branches. The first three branches each contain two $3\times3$ dilated convolutions, with each branch having a different dilation rate. The calculations for the first three branches are as Eq.\ref{eq:three}:
\begin{equation}
\begin{gathered}
D^{1}_{i}=cat(Conv1(F_{i}),Conv1(Conv1(F_{i})))\\
D^{2}_{i}=cat(Conv1(F_{i}),Conv3(Conv1(F_{i})))\\
D^{3}_{i}=cat(Conv1(F_{i}),Conv5(Conv1(F_{i})))
\label{eq:three}
\end{gathered}
\end{equation}
where $Convd(\cdot)$ ($d \in \{1, 3, 5\}$) denotes a $3\times3$ convolutional layer and Batch Normalization, $d$ is dilation rate. $F_{i}$ ($i \in \{1, 2, 3, 4\}$) comes from the encoder and ${D}^j_i$ ($j \in \{1, 2, 3\}$) is the result of the module's three feature mappings.

The final branch extracts high-frequency features using the BA to enhance attention to details. Subsequently, the features from all four branches are combined and the features are extracted through a convolutional layer to obtain $C_{i}$, as expressed in the Eq. \ref{eq:four} and Eq. \ref{eq:final}:
\begin{align}
\label{eq:four}
&D_{i}=BN(D^{1}_{i}+D^{2}_{i}+D^{3}_{i})\\
\label{eq:final}
&C_{i}=ReLU(Conv1(BA(F_{i})+D_{i}))
\end{align}
where $BN(\cdot)$ denotes Batch Normalization, $ReLU(\cdot)$ stands for Rectified Linear Unit. $BA(\cdot)$ represents the BA Block.

\begin{figure}  
\centerline{\includegraphics[width=\columnwidth]{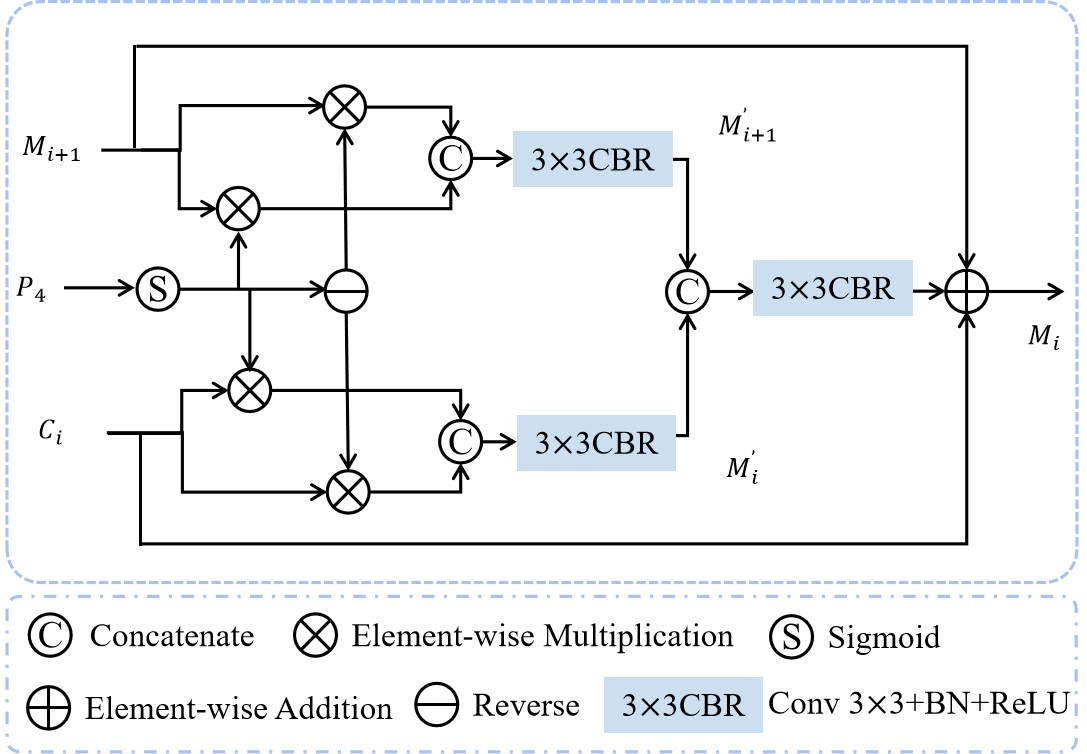}}
\caption{The details of Mask-guided Feature Aggregation Module (MFAM).}
\label{fig:MFAM}
\end{figure}

\subsubsection{Mask-guided feature aggregation module}\label{CC}

To achieve comprehensive multi-level feature fusion, we propose a Mask-guided Feature Aggregation Module (MFAM). As illustrated in Figure \ref{fig:MFAM}, this module effectively fuses features from top to bottom using masks generated by the Cascaded Mask Decoder (CMD). We extend $P_{4}$ to match the channel dimension of $C_{i}$ ($i \in \{1,2,3\}$) through broadcasting. We apply $P_{4}$ to the sigmoid function to obtain the foreground attention map $S_{4}$. Inspired by \cite{zhang2021bilateral}, we obtain the background attention map $B_{4}$ through the utilization of the inverse operation. The attention map is multiplied with the respective high-level and low-level features.
By concatenating the results, we obtain two edge enhancement maps $M'_{i+1}$ and $M'_{i}$. This process can be formulated as Eq.\ref{eq:sig}, Eq.\ref{eq:fa}, Eq.\ref{eq:up} and Eq.\ref{eq:down}:
\begin{align}
\label{eq:sig}
&S_{4}=\sigma(P_{4})\\
\label{eq:fa}
&B_{4}=E-\sigma(P_{4})\\
\label{eq:up}
&M'_{i+1}=Conv_{3}(cat(S_{4}\otimes M_{i+1},B_{4}\otimes M_{i+1}))\\
\label{eq:down}
&M'_{i}=Conv_{3}(cat(S_{4}\otimes C_{i},B_{4}\otimes C_{i}))
\end{align}
where $\sigma$ is the Sigmoid function, $E$ denotes the matrix with all entries equal to one and $\otimes$ denotes the element-wise multiplication. $Conv_{3}(\cdot)$ denotes a $3\times3$ convolution with batch normalization and the ReLU activation. $cat(\cdot)$ represents the concatenation operation. It is worth noting that when $i=3$, $M_{4}$ represents $C_{4}$.

After concatenating the high-level edge enhancement image $M'_{i+1}$ with the low-level edge enhancement image $M'_{i}$, element-wise summation of $M_{i+1}$ and $C_{i}$ is performed to retain the original important information. The above operations can be expressed as Eq.\ref{eq:Out}:
\begin{equation}
\begin{gathered}
M_{i}=Conv_{3}(cat(M'_{i+1},M'_{i}))+M_{i+1}+C_{i}
\label{eq:Out}
\end{gathered}
\end{equation}
The outputs $P_{i}$ can be obtained by processing $M_{i}$ with a $3\times3$ convolution operation.

\subsection{Loss function} 
Binary Cross-Entropy (BCE) loss is a widely used loss function in camouflaged object detection. However, BCE evaluates the loss of each pixel independently and lacks a global perspective. Wei et al. \cite{wei2020f3net} proposed a hybrid loss combining Weighted Binary Cross-Entropy (wBCE) and Weighted Intersection over Union (wIoU) losses. The loss output is represented as Eq.\ref{eq:loss1}:
\begin{equation}
L= L_{wBCE} + L_{wIoU}
\label{eq:loss1}
\end{equation}
We employ this hybrid loss to supervise the four outputs of the model, the total training loss for the proposed model can be represented as Eq.\ref{eq:lossTotal}:
\begin{equation}
L_{total}=\sum_{i=1}^{4} L(P_{i},G)
\label{eq:lossTotal}
\end{equation}
where $P_{i}$ ($i \in \{1,2,3,4\}$) represents the $i$-th predicted camouflaged map. In the weakly supervised object detection experiments, $G$ represents $M$ in Eq.\ref{eq:SAM} and the mask processed by RPS. In the fully supervised object detection experiments, $G$ represents the ground truth. $P_{1}$ is the final prediction.

\section{Experiments}
In this section, we introduce the dataset, evaluation metrics and experimental details. We compared weakly supervised and fully supervised COD methods using three commonly used COD datasets. To further demonstrate the general performance of the proposed method, we also tested in SOD with bounding-box supervision on four commonly used SOD datasets. Ablation experiments were conducted on the proposed strategy and individual components of MGNet to verify their effectiveness.

\begin{table*}  
\caption{Quantitative comparison with point supervision, scribble supervision and bounding-box supervision on COD. The best results are highlighted in \textbf{Bold}.}
\resizebox{\textwidth}{!}{
\begin{tabular}{|c|c|c|cccc|cccc|cccc|}
\hline
\multirow{2}{*}{\centering \textbf{Model}}&\multirow{2}{*}{\centering \textbf{Pub/Year}}&\multirow{2}{*}{\centering \textbf{Supervision}}&\multicolumn{4}{c|}{\textbf{CAMO-Test}}&\multicolumn{4}{c|}{\textbf{COD10K-Test}}&\multicolumn{4}{c|}{\textbf{NC4K}} \\
\cline{4-15} 
&&&\textbf{\textit{$S_\alpha\uparrow$}}& \textbf{\textit{$F _\beta\uparrow$}}& \textbf{\textit{$\mathcal{M}\downarrow$}}& \textbf{\textit{$E_\phi\uparrow$}}&\textbf{\textit{$S_\alpha\uparrow$}}& \textbf{\textit{$F _\beta\uparrow$}}& \textbf{\textit{$\mathcal{M}\downarrow$}}& \textbf{\textit{$E_\phi\uparrow$}}&\textbf{\textit{$S_\alpha\uparrow$}}& \textbf{\textit{$F _\beta\uparrow$}}& \textbf{\textit{$\mathcal{M}\downarrow$}}& \textbf{\textit{$E_\phi\uparrow$}} \\

\hline
WSSA \cite{zhang2020weakly}&CVPR$_{20}$&Point&0.649 & 0.607 & 0.148 & 0.652 & 0.642 & 0.509 & 0.087 & 0.733 & 0.743 & 0.688 & 0.104 & 0.756 \\
SCWS \cite{yu2021structure}&AAAI$_{21}$&Point &0.687 & 0.624 & 0.142 & 0.672 & 0.738 & 0.593 & 0.082 & 0.777 & 0.754 & 0.695 & 0.098 & 0.767 \\
TEL \cite{liang2022tree}& CVPR$_{22}$ &Point  &0.645 & 0.662 & 0.133 & 0.674 & 0.727 & 0.623 & 0.063 & 0.803 & 0.766 & 0.725 & 0.085 & 0.795 \\
SCOD \cite{he2023weakly}&AAAI$_{23}$&Point&0.663 & 0.629 & 0.137 & 0.688 & 0.711 & 0.607 & 0.060 & 0.802 & 0.758 & 0.744 & 0.080 & 0.796 \\
SAM \cite{kirillov2023segment}&ICCV$_{23}$&Point&0.677 & 0.649 & 0.123 & 0.693 & 0.765 & 0.694 & 0.069 & 0.796 & 0.776 & 0.728 & 0.082 & 0.786 \\
WS-SAM \cite{he2024weakly}&NeurIPS$_{23}$&Point&0.718 & 0.703 & 0.102 &0.757 & 0.790 & 0.698 &0.039 &0.856 & 0.813 & 0.801 &0.057 &0.859 \\
WSSA \cite{zhang2020weakly}&CVPR$_{20}$&Scribble&0.696&0.615&0.118&0.786&0.684&0.536&0.071&0.770&0.761&0.657&0.091&0.779 \\
SCWS \cite{yu2021structure}&AAAI$_{21}$ &Scribble&0.713 & 0.658 & 0.102 & 0.795 & 0.710 & 0.602 & 0.055 & 0.805 & 0.784 & 0.723 & 0.073 & 0.814 \\
TEL \cite{liang2022tree}& CVPR$_{22}$   &Scribble&0.717 & 0.681 & 0.104 & 0.797 & 0.724 & 0.633 & 0.057 & 0.826 & 0.782 & 0.754 & 0.075 & 0.832 \\
SCOD \cite{he2023weakly}&AAAI$_{23}$&Scribble&0.735 & 0.709 & 0.092 & 0.815 & 0.733 & 0.637 & 0.049 & 0.832 & 0.779 & 0.751 & 0.064 & 0.853 \\
SAM \cite{kirillov2023segment}&ICCV$_{23}$&Scribble&0.731 & 0.682 & 0.105 & 0.774 & 0.772 & 0.695 & 0.046 & 0.828 & 0.763& 0.747 & 0.071 & 0.832 \\
WS-SAM \cite{he2024weakly}&NeurIPS$_{23}$&Scribble&0.759 & 0.742 & 0.092 & 0.818 & 0.803 & 0.719 &0.038 & 0.878 & 0.829 & 0.802 & 0.052 & 0.886 \\
SAM-COD \cite{chen2024sam}&ECCV$_{24}$&Bounding-box&0.837&-&0.062&0.901&0.842&-&0.028&0.914&0.867&-&\textbf{0.037}&0.923\\
\hline
BoxSAM-P (Ours) &&Point&0.745 &0.757 & 0.102& 0.749 &0.808&0.756 &0.042 & 0.851 & 0.810 & 0.814 & 0.069 & 0.823\\
BoxSAM-S (Ours)&&Scribble&0.830 &0.789 &0.084 &0.860&0.836 &0.749 & 0.044 & 0.883&0.860 &0.825&0.052 &0.894\\

BoxSAM (Ours) &&Bounding-box&\textbf{0.859} & \textbf{0.842} & \textbf{0.057} & \textbf{0.908} & \textbf{0.857} & \textbf{0.789} & \textbf{0.027} & \textbf{0.919} & \textbf{0.877} & \textbf{0.854} & \textbf{0.037} & \textbf{0.925}\\
\hline
\end{tabular}}
\label{tab:compare1}
\end{table*}

\begin{figure*}  
\centerline{\includegraphics[width=1\textwidth]{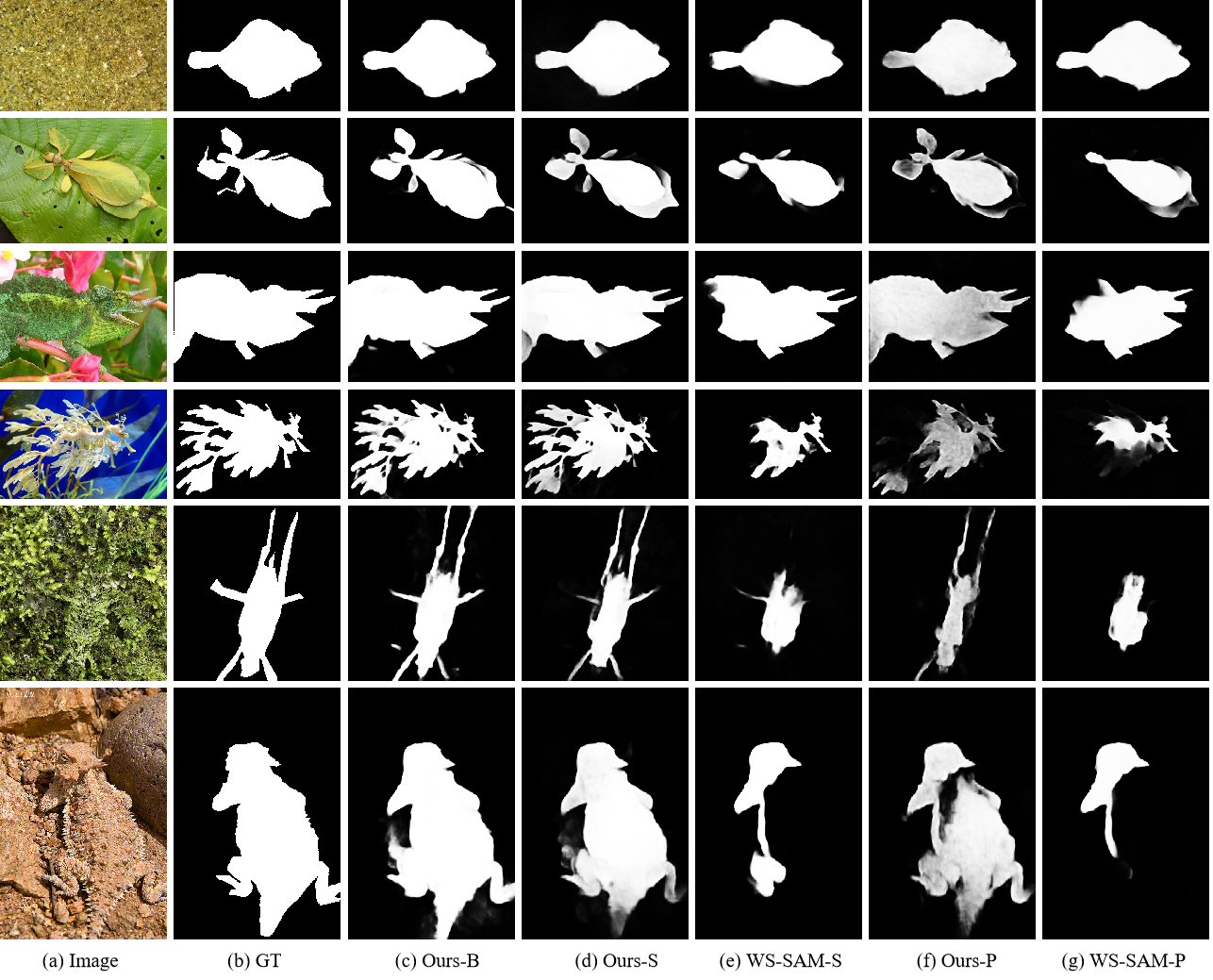}}
\caption{Qualitative comparison with WSCOD methods. (a) RGB images. (b) Ground truth. (c)-(g) The detection results obtained by (c) Our method based on bounding-box supervision, (d) Our method based on scribble supervision, (e) WS-SAM \cite{he2024weakly} based on scribble supervision, (f) Our method based on point supervision, (g) WS-SAM \cite{he2024weakly} based on point supervision.}
\label{fig:weakly}
\end{figure*}

\begin{table*}  
\caption{Quantitative comparison with 3 bounding-box supervision methods on SOD. The best results are highlighted in \textbf{Bold}.}
\setlength{\tabcolsep}{13pt} 
\resizebox{\textwidth}{!}{
\begin{tabular}{|c|c|cc|cc|cc|cc|}
\hline
\multirow{2}{*}{\centering \textbf{Model}}&\multirow{2}{*}{\centering \textbf{Pub/Year}}&\multicolumn{2}{c|}{\textbf{ECSSD}}&\multicolumn{2}{c|}{\textbf{HKU-IS}} &\multicolumn{2}{c|}{\textbf{DUT-OMRON}}&\multicolumn{2}{c|}{\textbf{DUTS-TE}} \\
\cline{3-10} 
&& \textbf{\textit{$F^{m} _\beta\uparrow$}}& \textbf{\textit{$\mathcal{M}\downarrow$}}& \textbf{\textit{$F^{m}  _\beta\uparrow$}}& \textbf{\textit{$\mathcal{M}\downarrow$}}& \textbf{\textit{$F^{m} _\beta\uparrow$}}& \textbf{\textit{$\mathcal{M}\downarrow$}}& \textbf{\textit{$F^{m} _\beta\uparrow$}}& \textbf{\textit{$\mathcal{M}\downarrow$}} \\
\hline
SBB \cite{liu2021weakly} &TCSVT$_{21}$&0.878 & 0.072 & 0.869 & 0.057 & 0.751 & 0.075& 0.775 & 0.073\\
Wang et al. \cite{wang2023weakly} &JOCA$_{23}$&0.894 & 0.062 & 0.880 & 0.052 & 0.791 & 0.065& 0.806 & 0.062 \\
Liu et al. \cite{Liu2024weakly}&ERA$_{24}$&0.945 & 0.035 &0.932 & 0.031 & 0.793 & 0.055& 0.889 & 0.036 \\
BoxSAM (Ours) &&\textbf{0.952} &\textbf{0.029}  & \textbf{0.933} & \textbf{0.029} & \textbf{0.825} & \textbf{0.048} & \textbf{0.897} & \textbf{0.033}\\
\hline
\end{tabular}}
\label{tab:sodcompare}
\end{table*}

\begin{table*}  
\caption{Quantitative comparison with 7 CNN-based methods and 10 Transformer-based methods on COD. The best results are highlighted in \textbf{Bold}.}
\resizebox{\textwidth}{!}{
\begin{tabular}{|c|c|c|cccc|cccc|cccc|}
\hline
\multirow{2}{*}{\centering \textbf{Model}}&\multirow{2}{*}{\centering \textbf{Pub/Year}}&\multirow{2}{*}{\centering \textbf{Backbone}}&\multicolumn{4}{c|}{\textbf{CAMO-Test}}&\multicolumn{4}{c|}{\textbf{COD10K-Test}}&\multicolumn{4}{c|}{\textbf{NC4K}} \\
\cline{4-15} 
&&&\textbf{\textit{$S_\alpha\uparrow$}}& \textbf{\textit{$F _\beta\uparrow$}}& \textbf{\textit{$\mathcal{M}\downarrow$}}& \textbf{\textit{$E_\phi\uparrow$}}&\textbf{\textit{$S_\alpha\uparrow$}}& \textbf{\textit{$F _\beta\uparrow$}}& \textbf{\textit{$\mathcal{M}\downarrow$}}& \textbf{\textit{$E_\phi\uparrow$}}&\textbf{\textit{$S_\alpha\uparrow$}}& \textbf{\textit{$F _\beta\uparrow$}}& \textbf{\textit{$\mathcal{M}\downarrow$}}& \textbf{\textit{$E_\phi\uparrow$}} \\
\hline
\multicolumn{15}{|c|}{\centering \textbf{CNN-Based Methods}}\\
\hline
SINet \cite{fan2020camouflaged}&CVPR$_{20}$&ResNet50&0.745 & 0.712 & 0.092 & 0.804 & 0.776 & 0.667 & 0.043 & 0.864 & 0.808 & 0.768 & 0.058 & 0.871\\
UGTR \cite{yang2021uncertainty}&ICCV$_{21}$&ResNet50&0.785 & 0.749 & 0.086 & 0.823 & 0.818 & 0.671 & 0.035 & 0.853 & 0.839 & 0.779 & 0.052 & 0.874\\
SINet-V2 \cite{fan2021concealed}&TPAMI$_{22}$&Res2Net50 &0.822 & 0.779 & 0.070 & \textbf{0.882} & 0.815 & 0.682 & 0.037 & 0.887 & 0.847 & 0.792 & 0.048 & 0.903 \\
BSANet \cite{zhu2022can}& AAAI$_{22}$ &Res2Net50  &0.794 & 0.768 & 0.079 & 0.851 & 0.818 & 0.723 & 0.034 & 0.891 & 0.841 & 0.805 & 0.048 & 0.897 \\
PFNet+ \cite{mei2023distraction}&SCIS$_{23}$&ResNet50 &0.791 & 0.764 & 0.080 & 0.850 & 0.806 & 0.698 & 0.037 & 0.884 & - & - & - & -\\
MRR-Net \cite{yan2023camouflaged}&TNNLS$_{23}$&Res2Net50 &0.826 & 0.793 & 0.070 & 0.880 & 0.835 & 0.726 & 0.032 & 0.901 & 0.857 & 0.807 & 0.044 & \textbf{0.906} \\
CamoFocus \cite{khan2024camofocus}& WACV$_{24}$ &ResNet50   &0.812 & \textbf{0.794} & 0.071 & - & 0.830 & \textbf{0.749} & 0.033 & - & 0.847 & 0.812 & \textbf{0.043} & - \\
MGNet-R (Ours)& &Res2Net50  &\textbf{0.828} & 0.791 & \textbf{0.069} & 0.879 & \textbf{0.843} & 0.747 & \textbf{0.030} & \textbf{0.905} & \textbf{0.859} & \textbf{0.818} & 0.044 & \textbf{0.906} \\
\hline
\multicolumn{15}{|c|}{\centering \textbf{Transformer-Based Methods}}\\
\hline
DTINet \cite{liu2022boosting}& ICPR$_{22}$&MiT &0.856 & 0.821 & 0.050 & 0.916 & 0.824 & 0.702 & 0.034 & 0.896 & 0.863 & 0.809 & 0.041 & 0.917 \\
FSPNet \cite{huang2023feature}&
CVPR$_{23}$&ViT&0.856 & 0.829 & 0.050 & 0.899 & 0.851 & 0.736 & 0.026 & 0.895 & 0.879 & 0.826 & 0.035 & 0.915\\
MSCAF-Net \cite{liu2023mscaf}& TCSVT$_{23}$&PVTv2\_B2&0.873 & 0.848 & 0.046 & 0.929 & 0.865 & 0.775 & 0.024 & 0.927 & 0.887 & 0.852 & 0.032 & 0.929\\
DGNet \cite{ji2023deep}&MIR$_{23}$&PVTv2\_B2&0.866 & 0.829 & 0.051 & 0.919 & 0.844 & 0.743 & 0.029 & 0.913 & 0.875 & 0.832 & 0.037 & 0.925\\
DCNet \cite{yue2023dual}&TCSVT$_{23}$&PVTv2\_B2&0.870 & - & 0.050 & 0.922 & 0.873 & - & 0.022 & 0.934 & - & - & - & -\\
HitNet \cite{hu2023high}& AAAI$_{23}$&PVTv2\_B2&0.849 & - & 0.055 & 0.906 & 0.871 & - & 0.023 & 0.935 & 0.875 & - & 0.037 & 0.926\\
FSPNet+ \cite{yang2024spatial}&arXiv$_{24}$&ViT &0.858 & 0.836 & 0.049 & 0.908 & 0.847 & 0.753 & 0.026 & 0.899 & 0.877 & 0.836 & 0.035 & 0.920\\
IPNet \cite{wang2024ipnet}&EAAI$_{24}$&PVTv2\_B2&0.864 & 0.846 & 0.047 & 0.924 & 0.850 & 0.785 & 0.026 & 0.922 & - & - & - & -\\
ICEG \cite{he2023strategic}&ICLR$_{24}$&Swin&0.867 & 0.855 & 0.044 & 0.926 & 0.857 & 0.782 & 0.024 & 0.930 & 0.879 & 0.855 & 0.034 & 0.932\\
CamoFormer-P \cite{yin2024camoformer}&TPAMI$_{24}$&PVTv2\_B4&0.872&0.853&0.046&0.929&0.869&0.794&0.023&0.932&0.892&0.863&0.030&0.939\\
MGNet (Ours)	     
&&PVTv2\_B2&0.882 & 0.859 & 0.044 & 0.933 & 0.878 & 0.800 & 0.022 & 0.934 & 0.893 & 0.860 & 0.032 & 0.936 \\
MGNet-B4 (Ours)     
&&PVTv2\_B4&\textbf{0.887} & \textbf{0.872} & \textbf{0.041} & \textbf{0.941} & \textbf{0.883} & \textbf{0.807} & \textbf{0.021} & \textbf{0.938} & \textbf{0.896} & \textbf{0.868} & \textbf{0.029} & \textbf{0.942}\\
\hline
\end{tabular}}
\label{tab:compare2}
\end{table*}

\subsection{Datasets and evaluation metrics}
\subsubsection{Datasets}
We evaluated our method on three commonly used datasets, including CAMO dataset \cite{le2019anabranch}, COD10K dataset \cite{fan2020camouflaged}, and NC4K dataset \cite{lv2021simultaneously}. The CAMO dataset consists of 1,000 training images and 250 testing images. The COD10K dataset is the largest camouflaged object dataset with 3,040 images for training and 2,026 images for testing. The NC4K dataset primarily consists of natural images, with its 4,121 images exclusively designated for use as the test set.

In camouflaged object detection task, there are few weakly supervised methods based on bounding-boxes. To verify the effectiveness of the proposed bounding-box based object detection method, we conducted experiments on SOD. We evaluated our method on four commonly used datasets, including the DUTS-TR dataset \cite{wang2017learning}, the ECSSD dataset \cite{yan2013hierarchical}, the DUT-OMRON dataset \cite{yang2013saliency} and the HKU-IS dataset \cite{li2016visual}. DUTS-TR dataset contains 10,553 images. The ECSSD dataset consists of 1,000 images. The HKU-IS dataset contains 4,447 images. The DUT-OMRON dataset contains 5,168 images, and the scenes in this dataset are complex. The DUTS-TE dataset contains 5,019 test images, which contain important scenes for SOD.

To ensure experimental fairness in COD, we used the same dataset partitioning method as existing models \cite{yue2023dual}. Specifically, 1,000 images from the CAMO dataset and 3,040 images from the COD10K dataset were allocated to the training set, while the remaining 250 images from the CAMO dataset, 2,026 images from the COD10K dataset, and the entire NC4K dataset served as the test set. For SOD, we followed established practices \cite{Liu2024weakly}, using the DUTS-TR dataset for training and the ECSSD, DUT-OMRON, and HKU-IS datasets for testing.

\begin{figure*}  
\centerline{\includegraphics[width=1\textwidth]{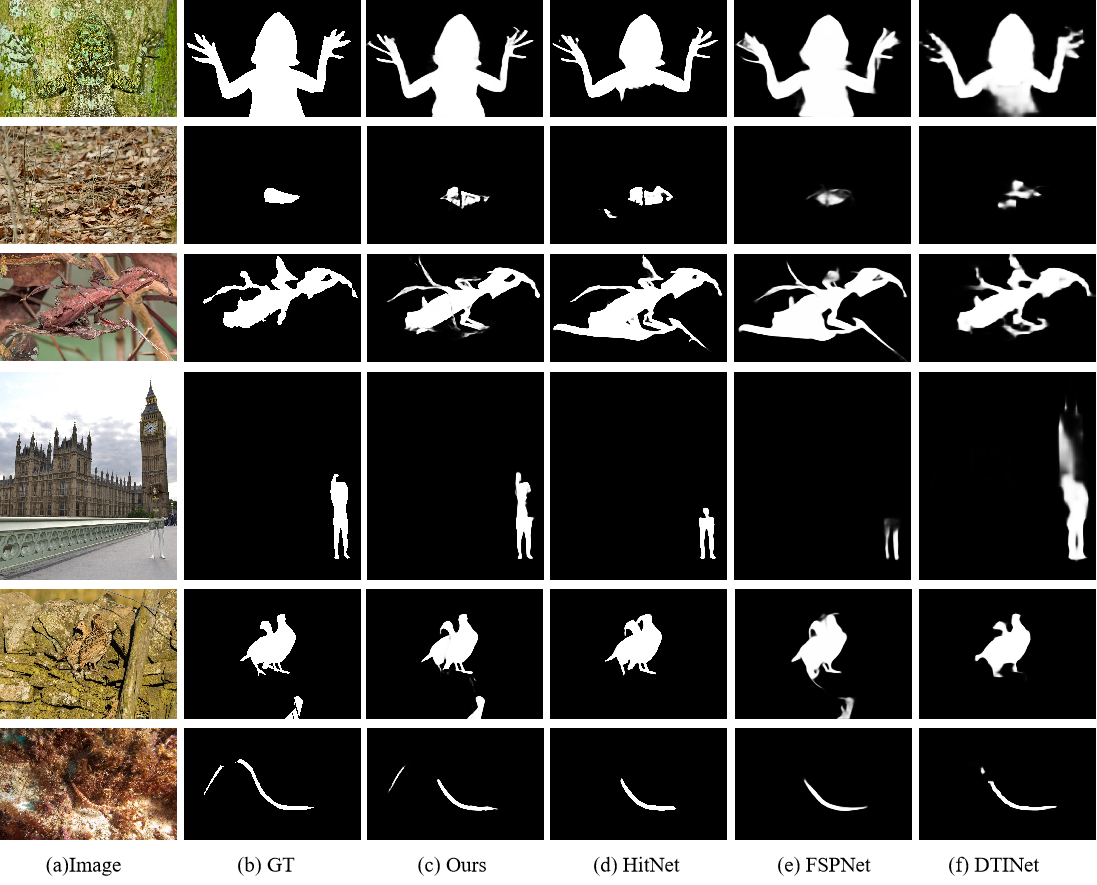}}
\caption{Qualitative comparison with Transformer-based COD methods. (a) RGB images. (b) Ground truth. (c)-(f) The detection results obtained by (c) Our method based on Transformer, (d) HitNet \cite{hu2023high}, (e) FSPNet \cite{huang2023feature}, (f) DTINet \cite{liu2022boosting}.}
\label{fig:transformer}
\end{figure*}

\subsubsection{Evaluation metrics}
To comprehensively compare our methods with competing COD methods, we chose four commonly used metrics, including the structural measure ($S_\alpha$), the adaptive F-measure ($F_\beta$), the mean absolute error ($\mathcal{M}$) and the mean E-measure ($E_\phi$). $S_\alpha$ is used to assess the structural similarity between the predicted map and the ground truth. $F$ takes into account both precision and recall. $\mathcal{M}$ represents the average value of the absolute error between the predicted value and the true value. $E$ is employed to assess segmentation results at both the pixel and image levels. Larger values of $S_\alpha$, $F_\beta$, $E_\phi$ and smaller values of $\mathcal{M}$ indicate better segmentation performance. 

To comprehensively compare our methods with competing SOD methods, we chose two commonly used metrics, including the mean absolute error ($\mathcal{M}$) and the maximum F-measure ($F^m_\beta$). Larger values of $F^m_\beta$ and smaller values of $\mathcal{M}$ indicate better segmentation performance.

\subsection{Implementation details}
Our proposed BoxSAM and MGNet are implemented based on PyTorch. For SAM, we adopt the ViT-H SAM model \cite{kirillov2023segment} to generate segmentation initial pseudo-labels. For MGNet, the backbone (i.e., PVTv2\_B2 \cite{wang2022pvt}, PVTv2\_B4 \cite{wang2022pvt} or Res2Net50 \cite{gao2019res2net}) is initialized with the parameters pre-trained on ImageNet large-scale dataset. In the weakly supervised training phase, the backbone employed is PVTv2\_B2, and the model is supervised by processed pseudo-labels during the training process. In contrast, during the fully supervised training phase, the model is supervised by ground truth labels. Apart from this, all other configurations of MGNet remain consistent. The input images are resized to $480\times480$. AdamW with an initial learning rate of 1e -- 4 is used as the optimizer, and a weight decay of 0.1 is used in the network training. The learning rate decays by a factor of 10 after 50 epochs. If either PVTv2\_B2 or Res2Net50 is chosen as the backbone, the batch size is set to 16. If PVTv2\_B4 is chosen as the backbone, the batch size is set to 12. The entire model is trained for 100 epochs. The whole model is trained and tested on a single NVIDIA 3090 GPU.

\begin{table*}  
\caption{Ablation study of BoxSAM on three datasets. The best results are highlighted in \textbf{Bold}.}
\setlength{\tabcolsep}{7pt} 
\resizebox{\textwidth}{!}{
\begin{tabular}{|c|cccc|cccc|cccc|}
\hline
\multirow{2}{*}{\centering \textbf{Model}}&\multicolumn{4}{c|}{\textbf{CAMO-Test}}&\multicolumn{4}{c|}{\textbf{COD10K-Test}}&\multicolumn{4}{c|}{\textbf{NC4K}} \\
\cline{2-13} 
&\textbf{\textit{$S_\alpha\uparrow$}}& \textbf{\textit{$F _\beta\uparrow$}}& \textbf{\textit{$\mathcal{M}\downarrow$}}& \textbf{\textit{$E_\phi\uparrow$}}&\textbf{\textit{$S_\alpha\uparrow$}}& \textbf{\textit{$F _\beta\uparrow$}}& \textbf{\textit{$\mathcal{M}\downarrow$}}& \textbf{\textit{$E_\phi\uparrow$}}&\textbf{\textit{$S_\alpha\uparrow$}}& \textbf{\textit{$F _\beta\uparrow$}}& \textbf{\textit{$\mathcal{M}\downarrow$}}& \textbf{\textit{$E_\phi\uparrow$}} \\
\hline
BoxSAM + w/o RPS&0.854 & 0.835 & 0.061 & 0.901 & \textbf{0.858} & 0.784 & 0.029 & 0.918 & 0.876 & 0.848 & 0.038 & 0.922 \\
MGNet + ADELE \cite{liu2022adaptive} &0.853 & 0.833 & 0.059 & 0.905 & 0.857 & 0.783 & 0.028 & 0.916 & \textbf{0.877} & 0.849 & 0.038 & 0.923 \\
BoxSAM (Ours)&\textbf{0.859} & \textbf{0.842} & \textbf{0.057} & \textbf{0.908} & 0.857 & \textbf{0.789} & \textbf{0.027} & \textbf{0.919} & \textbf{0.877} & \textbf{0.854} & \textbf{0.037} & \textbf{0.925}\\
\hline
\end{tabular}}
\label{tab:ablational1}
\end{table*}

\begin{table*}  
\caption{Ablation study of MGNet on two datasets. CMD: Cascaded Mask Decoder. CEM: Context Enhancement Module. MFAM: Mask-guided Feature Aggregation Module. The best results are highlighted in \textbf{Bold}.}
\setlength{\tabcolsep}{7pt} 
\resizebox{\textwidth}{!}{
\begin{tabular}{|c|c|ccc|cccc|cccc|}
\hline
\multirow{2}{*}{\centering \textbf{No}}&\multirow{2}{*}{\centering \textbf{Backbone}}&\multicolumn{3}{c|}{\textbf{Module}}&\multicolumn{4}{c|}{\textbf{CAMO-Test}}&\multicolumn{4}{c|}{\textbf{COD10K-Test}} \\
\cline{3-13} 
&&CMD&CEM&MFAM&\textbf{\textit{$S_\alpha\uparrow$}}& \textbf{\textit{$F _\beta\uparrow$}}& \textbf{\textit{$\mathcal{M}\downarrow$}}& \textbf{\textit{$E_\phi\uparrow$}}&\textbf{\textit{$S_\alpha\uparrow$}}& \textbf{\textit{$F _\beta\uparrow$}}& \textbf{\textit{$\mathcal{M}\downarrow$}}& \textbf{\textit{$E_\phi\uparrow$}} \\
\hline
\#1&PVTv2\_B2&&&&0.869&	0.838&	0.051&0.918	&0.866	&0.775	&0.024	&0.930 \\
\#2&PVTv2\_B2&\checkmark&&&0.875&0.843&	0.048	&0.923	&0.871	&0.778	&0.023	&0.931 \\
\#3&PVTv2\_B2&&\checkmark&&0.871&0.840	&0.049	&0.922	&0.874	&0.792	&0.023	&0.934 \\
\#4&PVTv2\_B2&\checkmark&\checkmark&&0.875	&0.845	&0.048	&0.925	&0.876	&0.795	&0.023	&0.934 \\
\#5&PVTv2\_B2&\checkmark&&\checkmark&0.876	&0.855	&0.046	&0.925	&0.876	&0.803	&0.023	&0.933 \\
\#6&PVTv2\_B2&&\checkmark&\checkmark&0.877&0.849&0.047&0.928&0.877&0.795&0.022&0.935\\
\#7&PVTv2\_B2&\checkmark&\checkmark&\checkmark&0.882	&0.859	&0.044	&0.933	&0.878	&0.800	&0.022	&0.934\\
\#8&Res2Net50&\checkmark&\checkmark&\checkmark&0.828 & 0.791 &0.069 & 0.879 & 0.843 & 0.747 &0.030 &0.905\\
\#9&PVTv2\_B4&\checkmark&\checkmark&\checkmark&\textbf{0.887}	&\textbf{0.872}&\textbf{0.041}&\textbf{0.941}&\textbf{0.883}&\textbf{0.807}&\textbf{0.021}&\textbf{0.938}\\
\hline
\end{tabular}}
\label{tab:ablational2}
\end{table*}

\subsection{Comparison with state-of-the-arts}
To verify the superiority of our method in the task of WSCOD, we compared with several state-of-the-art methods, including WSSA \cite{zhang2020weakly}, SCWS \cite{yu2021structure}, TEL \cite{liang2022tree}, SCOD \cite{he2023weakly}, SAM \cite{kirillov2023segment}, WS-SAM \cite{he2024weakly} and SAM-COD \cite{chen2024sam}. The backbone architectures of WSSA, SCWS, TEL, SCOD, SAM, WS-SAM, and SAM-COD are VGG16, ResNet50, ResNet101, ResNet50, ViT-H, ViT-H and ResNet50, and ViT-H and PVT-B4, respectively. In BoxSAM with scribble annotations, we used pixel-level weighted data generated by \cite{he2024weakly} without any additional screening. In BoxSAM with point annotations, following the previous weakly supervised segmentation method \cite{gao2022weakly}, one point was randomly selected from the foreground and one point was randomly selected from the background.

Due to the scarcity of weak supervision based on bounding-boxes in COD, we compared the proposed BoxSAM with methods in salient object detection task, including SBB \cite{liu2021weakly}, the method proposed by Wang et al. \cite{wang2023weakly} and the method proposed by Liu et al. \cite{Liu2024weakly}.

For fully supervised COD, we conducted a comparison with 17 main camouflaged object detection methods, which included 7 CNN-based COD models (SINet \cite{fan2020camouflaged}, UGTR \cite{yang2021uncertainty}, SINet-V2 \cite{fan2021concealed}, BSANet \cite{zhu2022can}, PFNet+  \cite{mei2023distraction}, MRR-Net \cite{yan2023camouflaged}, CamoFocus \cite{khan2024camofocus}) and 10 Transformer-based COD models (DTINet \cite{liu2022boosting}, FSPNet \cite{huang2023feature},  MSCAF-Net \cite{liu2023mscaf}, DGNet \cite{ji2023deep}, DCNet \cite{yue2023dual}, HitNet \cite{hu2023high}, FSPNet+ \cite{yang2024spatial}, IPNet \cite{wang2024ipnet}, ICEG \cite{he2023strategic}, CamoFormer-P \cite{yin2024camoformer}). All predictions made by the competitors were provided by the authors.

\begin{figure}  
\centerline{\includegraphics[width=\columnwidth]{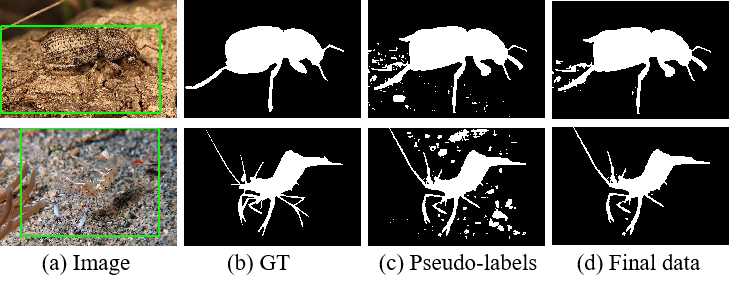}}
\caption{Some examples of masks. (a) RGB images. (b) Ground truth. (c) Some examples of SAM segmentation with bounding-boxes. (d) Some examples of masks after redundancy processing.}
\label{fig:RPS}
\end{figure}

\subsubsection{Quantitative evaluation}
Table \ref{tab:compare1} shows the quantitative comparison between BoxSAM and other weakly supervised COD models on public datasets. The results indicate that our model achieves superior performance on multiple evaluation metrics of the three datasets with scribble supervision and bounding-box supervision. In point supervision, BoxSAM improves the $S_\alpha$ by 3.8$\%$ and 2.3$\%$ on the CAMO and COD10K datasets compared with the suboptimal model, respectively. In scribble supervision, compared with the latest WSCOD method WS-SAM \cite{he2024weakly}, $S_\alpha$ and $F_\beta$ achieve an average enhancement of 5.7$\%$ and 4.5$\%$, respectively. In bounding-box supervision, our proposed method demonstrates a notable improvement in $S_\alpha$ metric, with an increase of 2.6$\%$, compared to the suboptimal method SAM-COD \cite{chen2024sam} on the CAMO dataset. Table \ref{tab:sodcompare} shows the quantitative comparison results of BoxSAM with other weakly supervised SOD models, BoxSAM improves $F^{m} _\beta$ by 4.0$\%$ on the DUT-OMRON dataset compared to the suboptimal model.

Table \ref{tab:compare2} shows the quantitative comparison results of MGNet with other CNN and transformer based COD models. The results indicate that our model achieves the best performance on $S_\alpha$ when using CNN as the backbone, outperforming other models that also use CNN backbones. In models employing PVTv2 as the backbone, on both the CAMO and COD10K datasets, the proposed MGNet demonstrates average improvements of 1.3\% and 2.3\% in the $S_\alpha$ and $F_\beta$ metrics, respectively, in comparison to the suboptimal method MSCAF-Net \cite{liu2023mscaf}. Moreover, compared to CamoFormer-P \cite{yin2024camoformer}, which uses PVTv2\_B4 as the backbone, our proposed MGNet\_B4 achieves improvements of 1.7$\%$ and 1.6$\%$ in $S_\alpha$ on the CAMO and COD10K datasets, respectively. 

\subsubsection{Qualitative comparisons}
Figure \ref{fig:weakly} compares the visual quality of WSCOD between our proposed BoxSAM method and WS-SAM using various supervised methods. This qualitative comparison covers various challenging scenarios, including highly intrinsic similarities (rows 1, 2 and 3), complex edges (row 4), and complex backgrounds (rows 5 and 6). Our method, which employs point and scribble supervision, effectively identifies more complete camouflaged objects. This improvement is largely attributed to the design of BoxSAM. Compared to point and scribble supervision, BoxSAM provides clearer edge details when using bounding-box supervision. This improvement is due to the use of the SAM model with bounding-box annotations as prompts.

Figure \ref{fig:transformer} presents the visualization of prediction results for Transformer-based fully supervised methods. This qualitative comparison covers various challenging scenarios scenes, including highly intrinsic similarities (row 1), tiny objects (rows 2 and 6), complex backgrounds (rows 3 and 5), and indefinable edges (row 4). Most methods struggle to extract edges of camouflaged objects, leading to incomplete segmentation (rows 1, 2, 3 and 4). In contrast, our method achieves complete segmentation of camouflaged objects. Additionally, small or obscured camouflaged objects are often missed (rows 5 and 6). MGNet delivers better prediction masks compared with other competing methods.

\begin{table*}  
\caption{Quantitative comparison with other models on BoxSAM. The best results are highlighted in \textbf{Bold}.}
\resizebox{\textwidth}{!}{
\begin{tabular}{|c|c|cccc|cccc|cccc|}
\hline
\multirow{2}{*}{\centering \textbf{Model}}&\multirow{2}{*}{\centering \textbf{Backbone}}&\multicolumn{4}{c|}{\textbf{CAMO-Test}}&\multicolumn{4}{c|}{\textbf{COD10K-Test}}&\multicolumn{4}{c|}{\textbf{NC4K}} \\
\cline{3-14} 
&&\textbf{\textit{$S_\alpha\uparrow$}}& \textbf{\textit{$F _\beta\uparrow$}}& \textbf{\textit{$\mathcal{M}\downarrow$}}& \textbf{\textit{$E_\phi\uparrow$}}&\textbf{\textit{$S_\alpha\uparrow$}}& \textbf{\textit{$F _\beta\uparrow$}}& \textbf{\textit{$\mathcal{M}\downarrow$}}& \textbf{\textit{$E_\phi\uparrow$}}&\textbf{\textit{$S_\alpha\uparrow$}}& \textbf{\textit{$F _\beta\uparrow$}}& \textbf{\textit{$\mathcal{M}\downarrow$}}& \textbf{\textit{$E_\phi\uparrow$}} \\
\hline
BoxSAM(MGNet$\to$SINet \cite{fan2020camouflaged})&ResNet50&0.728 & 0.704 & 0.104 & 0.753 & 0.763 & 0.621 & 0.048 & 0.814 & 0.803 & 0.755 & 0.063 & 0.842 \\

BoxSAM(MGNet$\to$SINet-V2\cite{fan2021concealed})&Res2Net50&0.799 & 0.766 & 0.084 & 0.852 & 0.802 & 0.680 & 0.041 & 0.874 & 0.836 & 0.788 & 0.052 & 0.893 \\

BoxSAM(MGNet$\to$HitNet \cite{hu2023high})&PVTv2\_B2&0.823&0.805&0.068&0.882&0.841&0.777&0.031&0.906&0.851&0.824&0.046&0.902 \\
BoxSAM(MGNet$\to$MSCAF-Net \cite{liu2023mscaf})&PVTv2\_B2&0.841 & 0.821 & 0.062 &0.897 & 0.845 & 0.763 &0.030 &0.912 & 0.873 & 0.845 &0.038 &0.923 \\
BoxSAM(Ours)&Res2Net50&0.806 & 0.786 & 0.077 & 0.856 & 0.824 & 0.735 & 0.034 & 0.890 & 0.844 & 0.813 & 0.049 & 0.893 \\

BoxSAM (Ours) &PVTv2\_B2&\textbf{0.859} & \textbf{0.842} & \textbf{0.057} & \textbf{0.908} & \textbf{0.857} & \textbf{0.789} & \textbf{0.027} & \textbf{0.919} & \textbf{0.877} & \textbf{0.854} & \textbf{0.037} & \textbf{0.925}\\
\hline
\end{tabular}}
\label{tab:compareBoxSAM}
\end{table*}

\subsection{Ablation study}
To demonstrate the effectiveness of our proposed modules, we conducted ablation studies on three benchmark datasets. The quantitative results of these studies are presented in Tables \ref{tab:ablational1}, \ref{tab:ablational2}, \ref{tab:compareBoxSAM}, \ref{tab:ablationalCMD} and \ref{tab:ablationalCEM}. 

\subsubsection{Effectiveness of RPS}
To reduce redundant information in the pseudo-labels generated by SAM, we develop the Redundancy Processing Strategy (RPS). To verify its effectiveness, we removed the RPS and used the pseudo-labels generated by SAM directly as the training set for the MGNet model. The results indicate that the performance of our proposed BoxSAM method decreases without the RPS strategy, as shown in Table \ref{tab:ablational1}. For instance, the $F_\beta$ drops by an average of 0.7\%. This demonstrates that the RPS enhances the quality of training labels for MGNet, thereby improving segmentation performance. Additionally, we compared RPS with ADELE \cite{liu2022adaptive}, and the results demonstrate that our method achieves superior segmentation performance. This is primarily because our method is specifically designed for pseudo-labels generated by SAM for camouflaged objects.
                                       
To enhance readability, we provide some demonstrations in Figure \ref{fig:RPS}. As shown in (c), the initial labels generated by SAM using bounding-boxes as prompts contain redundant information. Specifically, due to the considerable high similarity between objects and their surroundings, certain areas of the background are incorrectly identified as camouflaged objects. This incorrect identification adversely affects the model's accuracy and training effectiveness. In contrast, (d) demonstrates a significant improvement in the quality of the training labels after redundancy processing. By eliminating redundant information, this processing improves the quality of the training labels, enabling the MGNet to more accurately identify and segment camouflaged objects.

\begin{figure*}  
\centerline{\includegraphics[width=1\textwidth]{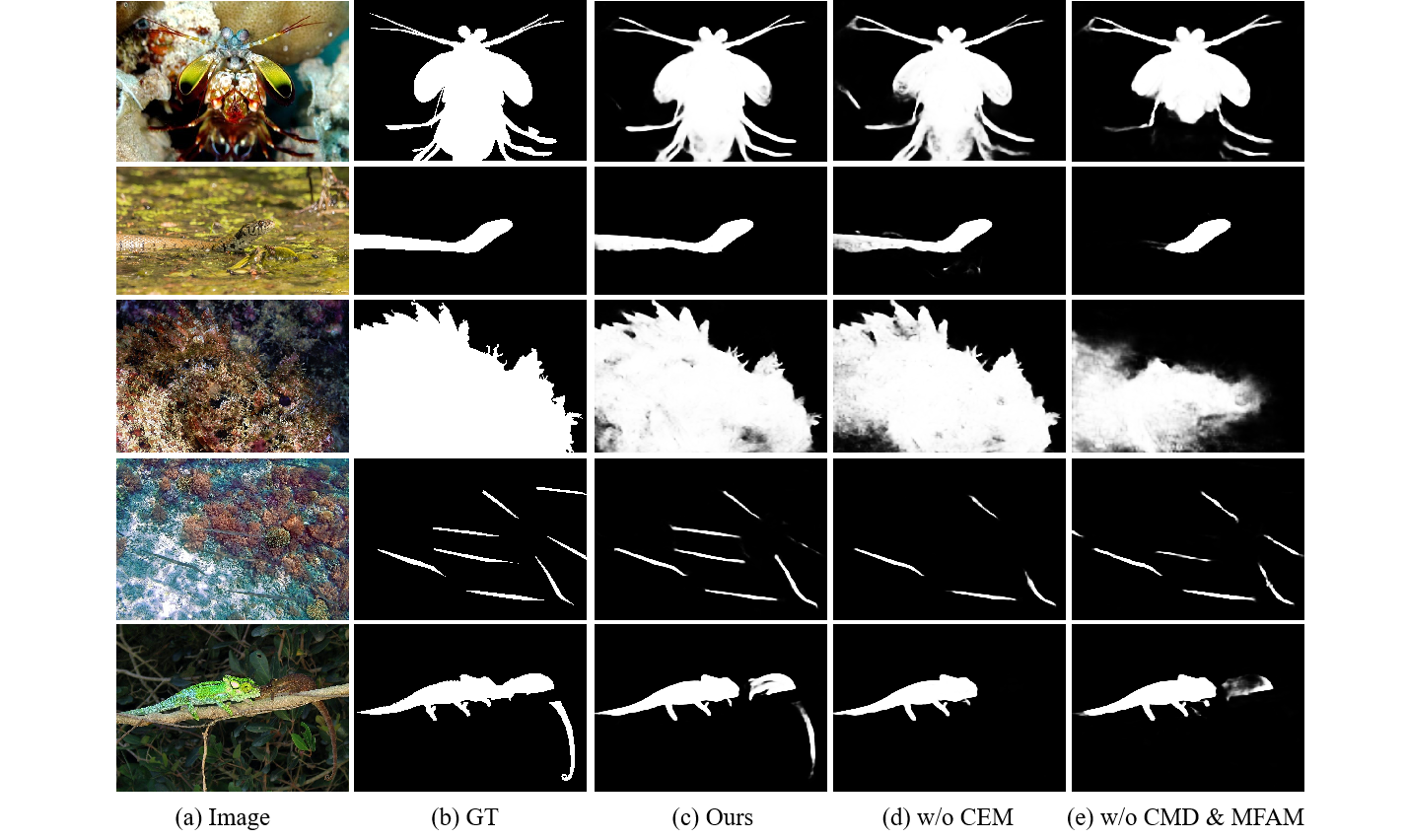}}
\caption{Visual verification of the effectiveness of the proposed
components. (a) RGB images. (b) Ground truth. (c)-(e) The detection results obtained by (c) Our method (MGNet), (d) MGNet without CEM, (e) MGNet without CMD and MFAM.}
\label{fig:ablationalImg}
\end{figure*} 

\begin{table*}  
\caption{Ablation study of Cascaded Mask Decoder on three datasets. The best result are highlighted in \textbf{Bold}.}

\resizebox{\textwidth}{!}{
\begin{tabular}{|c|cccc|cccc|cccc|}
\hline
\multirow{2}{*}{\centering \textbf{CEM}}&\multicolumn{4}{c|}{\textbf{CAMO-Test}}&\multicolumn{4}{c|}{\textbf{COD10K}}&\multicolumn{4}{c|}{\textbf{NC4K}} \\
\cline{2-13} 
&\textbf{\textit{$S_\alpha\uparrow$}}& \textbf{\textit{$F _\beta\uparrow$}}& \textbf{\textit{$\mathcal{M}\downarrow$}}& \textbf{\textit{$E_\phi\uparrow$}}&\textbf{\textit{$S_\alpha\uparrow$}}& \textbf{\textit{$F _\beta\uparrow$}}& \textbf{\textit{$\mathcal{M}\downarrow$}}& \textbf{\textit{$E_\phi\uparrow$}}&\textbf{\textit{$S_\alpha\uparrow$}}& \textbf{\textit{$F _\beta\uparrow$}}& \textbf{\textit{$\mathcal{M}\downarrow$}}& \textbf{\textit{$E_\phi\uparrow$}} \\
\hline
MGNet + w/o Cascade&0.877 & 0.852 & 0.046 & 0.931 &0.876&0.790&0.023&0.934&0.890 &0.858&\textbf{0.032}&0.935 \\
MGNet + w/o CBAM&0.879 & 0.856 & 0.046 & 0.929 &\textbf{0.878}&0.799&\textbf{0.022}&\textbf{0.936}&0.891 &0.859&\textbf{0.032}&0.935 \\
MGNet (Ours)&\textbf{0.882} & \textbf{0.859} &\textbf{0.044} & \textbf{0.933}&\textbf{0.878}	&\textbf{0.800}	&\textbf{0.022}	&0.934 &\textbf{0.893} & \textbf{0.860}& \textbf{0.032}&\textbf{0.936}\\
\hline
\end{tabular}}
\label{tab:ablationalCMD}
\end{table*}

\begin{table*}  
\caption{Ablation study of Context Enhancement Module on three datasets. The best results are highlighted in \textbf{Bold}.}

\resizebox{\textwidth}{!}{
\begin{tabular}{|c|cccc|cccc|cccc|}
\hline
\multirow{2}{*}{\centering \textbf{CEM}}&\multicolumn{4}{c|}{\textbf{CAMO-Test}}&\multicolumn{4}{c|}{\textbf{COD10K}}&\multicolumn{4}{c|}{\textbf{NC4K}} \\
\cline{2-13} 
&\textbf{\textit{$S_\alpha\uparrow$}}& \textbf{\textit{$F _\beta\uparrow$}}& \textbf{\textit{$\mathcal{M}\downarrow$}}& \textbf{\textit{$E_\phi\uparrow$}}&\textbf{\textit{$S_\alpha\uparrow$}}& \textbf{\textit{$F _\beta\uparrow$}}& \textbf{\textit{$\mathcal{M}\downarrow$}}& \textbf{\textit{$E_\phi\uparrow$}}&\textbf{\textit{$S_\alpha\uparrow$}}& \textbf{\textit{$F _\beta\uparrow$}}& \textbf{\textit{$\mathcal{M}\downarrow$}}& \textbf{\textit{$E_\phi\uparrow$}} \\
\hline
CEM $\to$ ASPP &0.879&0.852&0.046&0.928&0.877&0.796&0.023&\textbf{0.935}&0.892&0.856&0.032&\textbf{0.935} \\
MGNet + w/o Conv&0.878 & 0.856 & 0.045 & 0.926 &0.876&0.799&0.023&\textbf{0.935}& 0.891&\textbf{0.861}&0.033&0.933 \\
MGNet + w/o BA&0.880 & 0.857 & 0.046 & 0.929&0.877&0.795&0.023&0.933& 0.892&0.858&\textbf{0.032}&\textbf{0.937} \\
MGNet (Ours)&\textbf{0.882} & \textbf{0.859} & \textbf{0.044} & \textbf{0.933}&\textbf{0.878}	&\textbf{0.800}	&\textbf{0.022}	&0.934 &\textbf{0.893} & 0.860& \textbf{0.032}&0.936\\
\hline
\end{tabular}}
\label{tab:ablationalCEM}
\end{table*}

\begin{table*}  
\caption{Ablation study of loss on three datasets. The best result are highlighted in \textbf{Bold}.}

\resizebox{\textwidth}{!}{
\begin{tabular}{|c|cccc|cccc|cccc|}
\hline
\multirow{2}{*}{\centering \textbf{Loss}}&\multicolumn{4}{c|}{\textbf{CAMO-Test}}&\multicolumn{4}{c|}{\textbf{COD10K}}&\multicolumn{4}{c|}{\textbf{NC4K}} \\
\cline{2-13} 
&\textbf{\textit{$S_\alpha\uparrow$}}& \textbf{\textit{$F _\beta\uparrow$}}& \textbf{\textit{$\mathcal{M}\downarrow$}}& \textbf{\textit{$E_\phi\uparrow$}}&\textbf{\textit{$S_\alpha\uparrow$}}& \textbf{\textit{$F _\beta\uparrow$}}& \textbf{\textit{$\mathcal{M}\downarrow$}}& \textbf{\textit{$E_\phi\uparrow$}}&\textbf{\textit{$S_\alpha\uparrow$}}& \textbf{\textit{$F _\beta\uparrow$}}& \textbf{\textit{$\mathcal{M}\downarrow$}}& \textbf{\textit{$E_\phi\uparrow$}} \\
\hline
$L_{NC}$\cite{zhang2025learning}&0.845 & 0.836 & 0.060 & 0.903 &0.854&\textbf{0.792}&0.029&0.917&0.871 &0.853&0.038&0.923 \\
BoxSAM(Ours)&\textbf{0.859} & \textbf{0.842} & \textbf{0.057} & \textbf{0.908} & \textbf{0.857} & 0.789 & \textbf{0.027} & \textbf{0.919} & \textbf{0.877} & \textbf{0.854} & \textbf{0.037} & \textbf{0.925} \\
\hline
$L_{NC}$\cite{zhang2025learning}&0.870 & \textbf{0.865} & 0.045 & 0.932 &0.872&\textbf{0.813}&\textbf{0.022}&0.930&0.885 &\textbf{0.867}&\textbf{0.032}&0.932 \\
MGNet (Ours)&\textbf{0.882} & 0.859 &\textbf{0.044} & \textbf{0.933}&\textbf{0.878}	&0.800	&\textbf{0.022}	&\textbf{0.934} &\textbf{0.893} & 0.860& \textbf{0.032}&\textbf{0.936}\\
\hline
\end{tabular}}
\label{tab:ablationalLoss}
\end{table*}

\subsubsection{Effectiveness of MGNet}
To validate the effectiveness of MGNet in BoxSAM, we replaced MGNet in BoxSAM with other COD models for comparison. The models included SINet \cite{fan2020camouflaged}, SINet-V2 \cite{fan2021concealed}, HitNet \cite{hu2023high}, and MSCAF-Net \cite{liu2023mscaf}. The results are shown in Table \ref{tab:compareBoxSAM} . For instance, on the CAMO and COD10K datasets, the $S_\alpha$ improves by an average of 1.8\% compared to the suboptimal MSCAF-Net, and the $F _\beta$ improves by an average of 3.0\% compared to the suboptimal MSCAF-Net. The segmentation results in BoxSAM are better when using MGNet.

\subsubsection{Effectiveness of CMD}
We conducted ablation experiments to assess the effectiveness of the Cascaded Mask Decoder (CMD). The results of these experiments are illustrated in Tables \ref{tab:ablational2} and \ref{tab:ablationalCMD}. We contrasted the network with the cascade operation and CBAM \cite{woo2018cbam} removed (denoted as \#6) against the network with CMD included (denoted as \#7), as shown in Table \ref{tab:ablational2}. The removal of CMD results in performance drop, indicating that it contributes to improving detection performance.

In addition, we output the final prediction using the CMD (denoted as \#2), as shown in Table \ref{tab:ablational2}. The results demonstrate that our designed module improves upon the baseline metrics, indicating that the proposed module can generate more accurate initial masks and effectively guide the subsequent segmentation process. 

We further investigated this by removing the cascade operation and retaining the lowest-level features, as shown in the first row of Table \ref{tab:ablationalCMD}. This leads to a decline in segmentation performance, demonstrating that the cascade operation provides a more accurate mask for guiding subsequent generation. Additionally, we utilize CBAM to capture the details of camouflaged objects from different dimensions of the low-level feature map. To assess its effectiveness, we also examined the performance impact of removing CBAM from MGNet. As shown in the second row of Table \ref{tab:ablationalCMD}, the removal of CBAM causes a decline in performance. 

\subsubsection{Effectiveness of CEM}
To expand the receptive field and introduce rich context information, we propose the Context Enhancement Module (CEM). To evaluate the effectiveness of the CEM, we removed it (denoted as \#5) and the output of the encoder was directly input into the Mask-guided Feature Aggregation Module (MFAM) for aggregation, as shown in Table \ref{tab:ablational2}. By comparing the results in the \#5 and \#7 of Table \ref{tab:ablational2}, the removal of the CEM has resulted in performance drop. The quantitative comparison demonstrates that CEM can expand the receptive field and introduce rich contextual information. The efficacy of our CEM is visually demonstrated in Figure \ref{fig:ablationalImg}, as shown in (c) and (d), methods with the CEM can reduce the missing detection.

In addition, we incorporated the CEM (denoted as \#3) into the baseline model (denoted as \#1) to validate its effectiveness. Comparing \#3 with the baseline, we observe improved performance in all evaluation metrics. Notably, the $F_\beta$
shows an improvement of 2.1\% on the COD10K dataset. These experimental results demonstrate that our CEM significantly contributes to more accurate detection of camouflaged objects within the network.

Furthermore, we replaced CEM with ASPP \cite{chen2017deeplab}, as shown in the first row of Table \ref{tab:ablationalCEM}. The experimental results demonstrate that the CEM designed by us exhibits superior performance. We further investigated this by removing the dilated convolution, which leads to decreased segmentation performance, as shown in the second row of Table \ref{tab:ablationalCEM}. Additionally, we examined the impact of the BA \cite{hu2023high} on segmentation performance. As indicated in the third row of Table \ref{tab:ablationalCEM}, the removal of the BA resulted in performance degradation, hindering the accurate segmentation of detailed information.

\subsubsection{Effectiveness of MFAM}
 In order to effectively aggregate features, we design the Mask-guided Feature Aggregation Module (MFAM). To evaluate the effectiveness of the MFAM, we replaced it (denoted as \#4) with element-wise multiplication, convolutional layers and concatenation, as shown in Table \ref{tab:ablational2}. By comparing the results in the \#4 and \#7 of Table \ref{tab:ablational2}, we can observe that the removal of MFAM can cause a performance drop on CAMO and COD10K datasets. This indicates that MFAM plays a positive role in the COD task.

We further investigated the detection performance without CMD and MFAM (denoted as \#3), as shown in Table \ref{tab:ablational2}. By comparing the results in the \#3 and \#7 of Table \ref{tab:ablational2}, there is performance drop when removing CMD and MFAM from the proposed MGNet. For instance, there are $S_\alpha$ and $F_\beta$ drop of 1.3\% and 2.3\% on CAMO dataset, respectively. These findings suggest that feature aggregation based on mask guidance significantly enhances segmentation performance. The efficacy of our CMD and MFAM is demonstrated visually in Figure \ref{fig:ablationalImg}, as shown in (c) and (e), methods with CMD and MFAM can generate the predictions with clear edges.

\begin{table*}  
\caption{Quantitative comparison with 3 polyp segmentation methods and 2 COD methods. The best results are highlighted in \textbf{Bold}.}
\resizebox{\textwidth}{!}{
\begin{tabular}{|c|c|ccccc|ccccc|}
\hline
\multirow{2}{*}{\centering \textbf{Model}}&\multirow{2}{*}{\centering \textbf{Pub/Year}}&\multicolumn{5}{c|}{\textbf{CVC-ColonDB}}&\multicolumn{5}{c|}{\textbf{CVC-ClinicDB}} \\
\cline{3-12} 
&&\textbf{\textit{$mDice\uparrow$}}&\textbf{\textit{$mIoU\uparrow$}}&\textbf{\textit{$S_\alpha\uparrow$}}& \textbf{\textit{$\mathcal{M}\downarrow$}}&\textbf{\textit{$E_\phi\uparrow$}}&\textbf{\textit{$mDice\uparrow$}}&\textbf{\textit{$mIoU\uparrow$}}&\textbf{\textit{$S_\alpha\uparrow$}}& \textbf{\textit{$\mathcal{M}\downarrow$}}&\textbf{\textit{$E_\phi\uparrow$}}
\\
\hline
PraNet \cite{fan2020pranet}&MICCAI$_{20}$&0.703 & 0.634 & 0.815 & 0.037 & 0.835 & 0.916 & 0.870 & 0.946 & 0.007 & 0.973 
 \\
SSFormer \cite{shi2022ssformer}&MMSP$_{22}$&0.756 & 0.683 & 0.841 & 0.034 & 0.875 & 0.908 & 0.861 & 0.938 & 0.009 & 0.965  
 \\
 SINet-V2 \cite{fan2021concealed}&TPAMI$_{22}$&0.746 & 0.668 & 0.840 & 0.035 & 0.871 & 0.903 & 0.852 & 0.935 & 0.012 & 0.959 
 \\
 HitNet \cite{hu2023high}&AAAI$_{23}$&0.752 & 0.669 & 0.833 & 0.035 & 0.874 & 0.908 & 0.855 & 0.936 & 0.010 & 0.964 \\
 PVT-CASCADE \cite{rahman2023medical}&WACV$_{23}$&0.801&0.725&0.862&\textbf{0.029}&0.903&0.940&\textbf{0.895}&0.952&\textbf{0.006}&0.983\\
MGNet (Ours)&&\textbf{0.807} & \textbf{0.731} & \textbf{0.871} & 0.031 & \textbf{0.907} & \textbf{0.941} & \textbf{0.895} & \textbf{0.955} & \textbf{0.006} & \textbf{0.986} \\
\hline
\end{tabular}}
\label{tab:xr}
\end{table*}

\subsubsection{Effectiveness of PVTv2}
To effectively extract features, we choose the Pyramid Vision Transformer (PVT) \cite{wang2021pyramid} as our feature extraction module. To evaluate the effectiveness of the PVTv2 as the backbone, we replaced PVTv2\_B2 \cite{wang2022pvt} with Res2Net50 \cite{gao2019res2net}, which is pre-trained on the ImageNet large-scale dataset. By comparing the results in the \#7 and \#8 of Table \ref{tab:ablational2}, we can observe that Transformer-based model shows more obvious advantages compared with CNN-based model. Specifically, compared to methods using Res2Net50 as the backbone, $S_\alpha$, $F_\beta$, and $E_\phi$ achieved average improvements of 5.3\%, 7.8\%, and 4.7\%, respectively, on the CAMO and COD10K datasets.

Furthermore, we employed PVTv2\_B4 as the backbone for our model, as shown in \#9 of Table \ref{tab:ablational2}.  We can observe that PVTv2\_B4 achieves better segmentation performance compared to PVTv2\_B2. However, it is noteworthy that PVTv2\_B4 has a larger number of parameters than PVTv2\_B2 \cite{wang2022pvt}. Specifically, with the replacement of the feature extractor, the parameter count of MGNet with PVTv2\_B4 as the backbone reaches 63.67M, which is more than double that of MGNet with PVTv2\_B2 as the backbone.

\begin{figure}
\centerline{\includegraphics[width=\columnwidth]{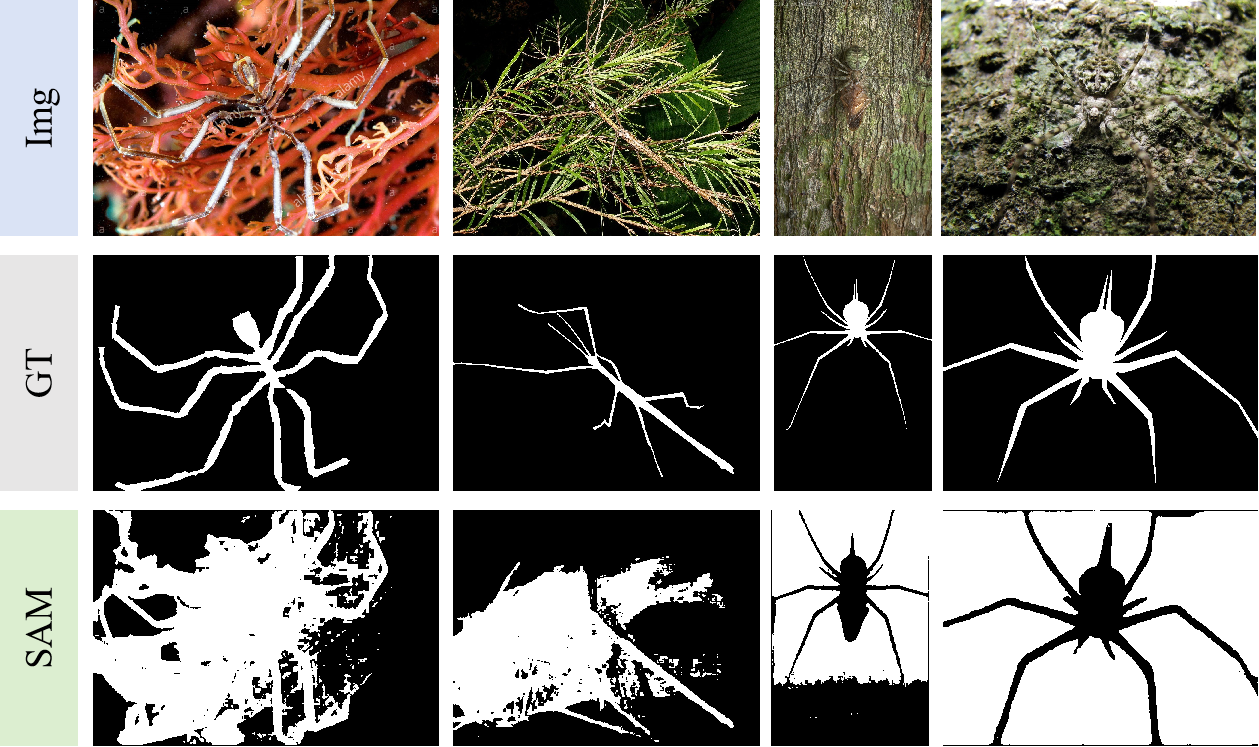}}
\caption{Failure cases of SAM segmentation results using bounding-box prompts.}
\label{fig:failure}
\end{figure}

\subsubsection{Effectiveness of Loss}
We evaluated the impact of the adopted loss function, as defined in Eq. \ref{eq:lossTotal}, and the noise correction loss ($L_{NC}$) \cite{zhang2025learning} on the experimental results. Following the settings in \cite{zhang2025learning}, we applied $L_{NC}$ as the loss function for both weakly supervised COD and fully supervised COD, as shown in Table \ref{tab:ablationalLoss}. The results show that $L_{NC}$ improves the $F _\beta$ metric, as it is designed to enhance the model's robustness to noisy labels \cite{zhang2025learning}. However, Eq. \ref{eq:lossTotal} demonstrates superior performance across the majority of the evaluated metrics, as it accounts for pixel-wise differences more effectively.

\subsubsection{Efficiency analysis}
To comprehensively evaluate our model, we investigated and compared our proposed MGNet with competing methods in terms of the floating point operations (FLOPs) and the number of parameters (Params) with three other COD models, as shown in Table \ref{tab:eff}. It can be seen that our proposed MGNet has lower computational complexity and fewer parameters. 

\begin{table}
    \centering
    \caption{Comparison of model parameters (Params) and FLOPs. }
    \resizebox{\columnwidth}{!}{
        \begin{tabular}{|c|c|c|c|c|}
            \hline
            \textbf{Model} & \textbf{MGNet (Ours)} & \textbf{DCNet} & \textbf{HitNet} & \textbf{MSCAF-Net} \\
            \hline
            Params & 26.48 M & 54.43 M & 25.73 M & 29.70 M \\
            FLOPs & 31.81 G & 94.74 G & 33.86 G & 30.04 G \\
            \hline
        \end{tabular}
    }
    \label{tab:eff}
\end{table}

\subsection{Limitation}
Although the BoxSAM proposed in this paper demonstrates positive effects in various experiments, the segmentation results generated by SAM with bounding-boxes as prompts exhibit failures, such as segmentation errors (columns 1 and 2) and failure to identify camouflaged objects (columns 3 and 4), as shown in Figure \ref{fig:failure}. Several studies \cite{ji2304sam, tang2023can, ji2024segment} have also highlighted the limitations of SAM in COD. Unfortunately, the approach outlined in this paper is not fully effective in addressing these challenges. Further research needs to focus on improving the accuracy of the generated initial pseudo-labels, such as by effectively combining multiple annotation methods as prompts for large models.

\subsection{COD-related applications}
In this study, we further validated the effectiveness of our proposed MGNet on two COD-related tasks. The application of polyp segmentation is detailed in Section \ref{PS}, while the application of defect detection is discussed in Section \ref{De}.

\subsubsection{Polyp segmentation} \label{PS}
Polyp segmentation is critical for the precise identification of early polyps, which plays a vital role in the clinical prevention of rectal cancer \cite{fan2020pranet}.
To validate the effectiveness of our method in polyp segmentation, we retrained MGNet following \cite{fan2020pranet}. We used 900 images from Kvasir \cite{jha2020kvasir} and 550 images from CVC-ClinicDB \cite{bernal2015wm} for the training set, while the remaining images from CVC-ClinicDB and CVC-ColonDB \cite{tajbakhsh2015automated} were used for testing. The evaluation metrics include $mDice$, $mIoU$, $S_\alpha$, $\mathcal{M}$, and $E_\phi$. Among these metrics, $Dice$ and $IoU$ are region-level similarity measures that primarily assess the internal consistency of object segmentation. Larger values of $mDice$, $mIoU$, $S_\alpha$, $E_\phi$ and smaller values of $\mathcal{M}$ indicate better segmentation performance.

\begin{figure}  
\centerline{\includegraphics[width=\columnwidth]{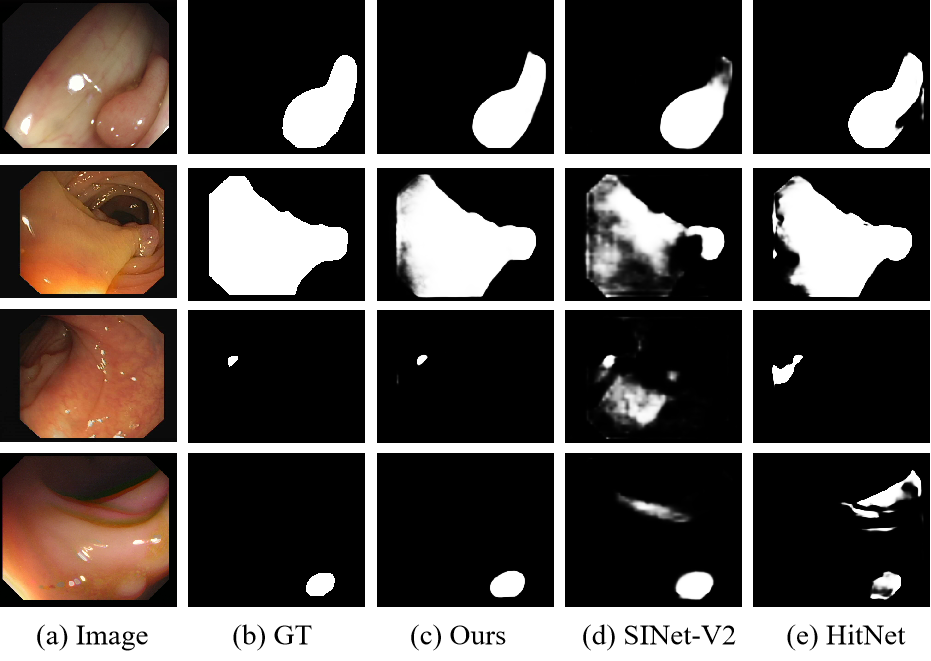}}
\caption{Qualitative comparison on CVC-ClinicDB  dataset \cite{bernal2015wm} and CVC-ColonDB \cite{tajbakhsh2015automated} dataset. (a) RGB images. (b) Ground truth. (c)-(e) The detection results obtained by (c) Our method, (d) SINet-V2\cite{fan2021concealed}, (e) HitNet\cite{hu2023high}. }
\label{fig:xr}
\end{figure}

We compared the performance of MGNet with three polyp segmentation methods, PraNet \cite{fan2020pranet}, SSFormer \cite{shi2022ssformer}, and PVT-CASCADE \cite{rahman2023medical}, and with two COD methods, SINet-V2 \cite{fan2021concealed} and HitNet \cite{hu2023high}. Some results were sourced from the original papers, while others were obtained by training and testing on equipment with the same configuration as MGNet. 
The quantitative results are presented in Table \ref{tab:xr}. MGNet outperforms PraNet \cite{fan2020pranet}, SSFormer \cite{shi2022ssformer}, SINet-V2 \cite{fan2021concealed}, and HitNet \cite{hu2023high} in terms of performance on CVC-ClinicDB. MGNet improves $mDice$, $mIoU$ and $S_\alpha$ by 0.7$\%$, 0.8$\%$ and 1.0$\%$, respectively, in CVC-ColonDB compared to the suboptimal model PVT-CASCADE \cite{rahman2023medical}. The visualization results in Figure \ref{fig:xr} illustrate the segmentation performance of MGNet in comparison to SINet-V2 \cite{fan2021concealed} and HitNet \cite{hu2023high}. This qualitative comparison covers large polyps (rows 1 and 2) and tiny polyps (rows 3 and 4). As shown in Figure \ref{fig:xr}, our proposed MGNet can identify polyps well and perform better than competing methods in these challenging cases.

\subsubsection{Defect detection} \label{De}
Defect detection is the process of identifying and locating defects or anomalies in products or materials through various techniques and methods, with the aim of ensuring that products meet quality standards and can be used safely. To validate our method’s effectiveness in defect detection, we retrained MGNet on the CDS2K dataset \cite{fan2023advances}, using 80\% of the dataset for training and the remaining 20\% for testing. The evaluation metrics include $S_\alpha$, $F_\beta$, $\mathcal{M}$ and $E_\phi$.

\begin{table}  
\caption{Quantitative comparison with 3 COD methods on the CDS2K dataset. The best results are highlighted in \textbf{Bold}.}
\resizebox{\columnwidth}{!}{
\begin{tabular}{|c|c|cccc|}
\hline
\multirow{2}{*}{\centering \textbf{Model}}&\multirow{2}{*}{\centering \textbf{Pub/Year}}&\multicolumn{4}{c|}{\textbf{CDS2K}}\\
\cline{3-6} 
&&\textbf{\textit{$S_\alpha\uparrow$}}& \textbf{\textit{$F _\beta\uparrow$}}& \textbf{\textit{$\mathcal{M}\downarrow$}}& \textbf{\textit{$E_\phi\uparrow$}}
\\
\hline
SINet \cite{shi2022ssformer}&CVPR$_{20}$&0.623 & 0.421 & 0.061 & 0.749 
 \\
 SINet-V2 \cite{fan2021concealed}&TPAMI$_{22}$&0.832 & 0.671 & 0.010 & 0.906 
 \\
MSCAF-Net \cite{hu2023high}&TCSVT$_{23}$&0.865 & 0.719 & \textbf{0.009} & 0.933 \\
MGNet (Ours)&&\textbf{0.877} & \textbf{0.772} & \textbf{0.009} & \textbf{0.944} \\
\hline
\end{tabular}}
\label{tab:dd}
\end{table}

\begin{figure}  
\centerline{\includegraphics[width=\columnwidth]{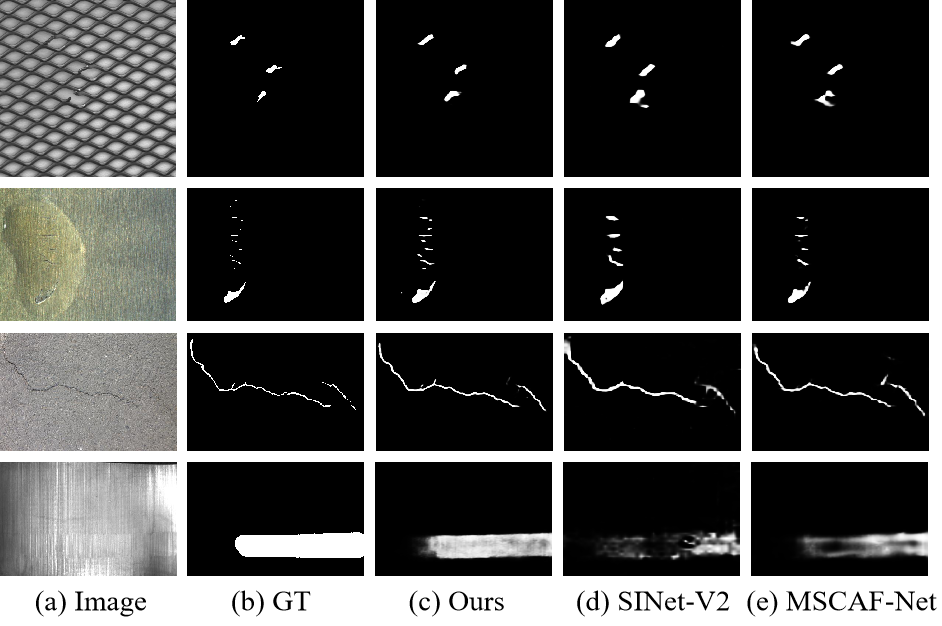}}
\caption{Qualitative comparison on CDS2K dataset \cite{fan2023advances}. (a) RGB images. (b) Ground truth. (c)-(e) The detection results obtained by (c) Our method, (d) SINet-V2 \cite{fan2021concealed}, (e) MSCAF-Net \cite{liu2023mscaf}. }
\label{fig:dd}
\end{figure}

We compared the performance of MGNet with three COD methods, SINet \cite{fan2020camouflaged}, SINet-V2 \cite{fan2021concealed} and MSCAF-Net \cite{liu2023mscaf}. The results were obtained by training and testing on equipment with the same configuration as MGNet. The quantitative results are presented in Table \ref{tab:dd}. MGNet improves $S_\alpha$, $F_\beta$, and $E_\phi$ by 1.4$\%$, 7.3$\%$, and 1.2$\%$, respectively, compared to the suboptimal model MSCAF-Net \cite{liu2023mscaf}. The visualization results in Figure \ref{fig:dd} illustrate the segmentation performance of MGNet in comparison to SINet-V2 \cite{fan2021concealed} and MSCAF-Net \cite{liu2023mscaf}. This qualitative comparison covers texture-grid (row 1), steel surface (row 2), road surface (row 3) and magnetic tile surface (row 4). As shown in Figure \ref{fig:dd}, our proposed MGNet can effectively identify defects across different materials and outperforms competing methods in these challenging cases.

\section{Discussions and conclusion}
In this paper, we propose a new WSCOD method characterized by two key components. The first component generates pseudo-labels by using the bounding-boxes of camouflaged objects as prompts for the SAM model.The second component involves redundant processing strategy that provides high-quality pixel-level pseudo-labels for training our designed Mask-guided Network (MGNet). MGNet includes a Cascaded Mask Decoder (CMD) to progressively integrate information from each layer to produce masks. The Context Enhancement Module (CEM) supplements the missing details of features and avoids the missing detection. Finally, the Mask-guided Feature Aggregation Module (MFAM) aggregates features across different levels guided by masks to generate predictions with clear object edges.

BoxSAM is evaluated on three widely used COD datasets, demonstrating superior performance compared to current state-of-the-art WSCOD methods. It also achieves state-of-the-art results in weakly supervised SOD using bounding-box supervision. MGNet performs effectively in both COD and other COD-related applications, such as polyp segmentation and defect detection. However, BoxSAM still exhibits some limitations in segmenting camouflaged objects. Further research needs to focus on improving the accuracy of the generated initial pseudo-labels, such as by effectively combining multiple annotation methods as prompts for large models.
\bibliographystyle{cas-model2-names}

\bibliography{cas-refs}

@inproceedings{fan2020camouflaged,
  title={Camouflaged object detection},
  author={Fan, Deng-Ping and Ji, Ge-Peng and Sun, Guolei and Cheng, Ming-Ming and Shen, Jianbing and Shao, Ling},
  booktitle={Proceedings of the IEEE/CVF conference on computer vision and pattern recognition},
  pages={2777--2787},
  year={2020}
}

@article{fan2023advances,
  title={Advances in deep concealed scene understanding},
  author={Fan, Deng-Ping and Ji, Ge-Peng and Xu, Peng and Cheng, Ming-Ming and Sakaridis, Christos and Van Gool, Luc},
  journal={Visual Intelligence},
  volume={1},
  number={1},
  pages={16},
  year={2023},
  publisher={Springer}
}

@article{liu2024weakly,
  title={Weakly supervised salient object detection via bounding-box annotation and SAM model},
  author={Liu, Xiangquan and Huang, Xiaoming},
  journal={Electronic Research Archive},
  volume={32},
  number={3},
  pages={1624--1645},
  year={2024}
}

@inproceedings{fan2020pranet,
  title={Pranet: Parallel reverse attention network for polyp segmentation},
  author={Fan, Deng-Ping and Ji, Ge-Peng and Zhou, Tao and Chen, Geng and Fu, Huazhu and Shen, Jianbing and Shao, Ling},
  booktitle={International conference on medical image computing and computer-assisted intervention},
  pages={263--273},
  year={2020},
  organization={Springer}
}

@article{fuentes2017robust,
  title={A robust deep-learning-based detector for real-time tomato plant diseases and pests recognition},
  author={Fuentes, Alvaro and Yoon, Sook and Kim, Sang Cheol and Park, Dong Sun},
  journal={Sensors},
  volume={17},
  number={9},
  pages={2022},
  year={2017},
  publisher={MDPI}
}

@inproceedings{song2010new,
  title={A new camouflage texture evaluation method based on WSSIM and nature image features},
  author={Song, Liming and Geng, Weidong},
  booktitle={2010 International conference on multimedia technology},
  pages={1--4},
  year={2010},
  organization={IEEE}
}

@article{pan2011study,
  title={Study on the camouflaged target detection method based on 3D convexity},
  author={Pan, Yuxin and Chen, Yiwang and Fu, Qiang and Zhang, Ping and Xu, Xin and others},
  journal={Modern Applied Science},
  volume={5},
  number={4},
  pages={152},
  year={2011},
  publisher={Canadian Center of Science and Education}
}

@article{hou2011detection,
  title={Detection of the mobile object with camouflage color under dynamic background based on optical flow},
  author={Hou, Jianqin Yin Yanbin Han Wendi and Li, Jinping},
  journal={Procedia Engineering},
  volume={15},
  pages={2201--2205},
  year={2011},
  publisher={Elsevier}
}

@inproceedings{bearman2016s,
  title={What’s the point: Semantic segmentation with point supervision},
  author={Bearman, Amy and Russakovsky, Olga and Ferrari, Vittorio and Fei-Fei, Li},
  booktitle={European conference on computer vision},
  pages={549--565},
  year={2016},
  organization={Springer}
}

@inproceedings{huang2023feature,
  title={Feature shrinkage pyramid for camouflaged object detection with transformers},
  author={Huang, Zhou and Dai, Hang and Xiang, Tian-Zhu and Wang, Shuo and Chen, Huai-Xin and Qin, Jie and Xiong, Huan},
  booktitle={Proceedings of the IEEE/CVF conference on computer vision and pattern recognition},
  pages={5557--5566},
  year={2023}
}

@inproceedings{kirillov2023segment,
  title={Segment anything},
  author={Kirillov, Alexander and Mintun, Eric and Ravi, Nikhila and Mao, Hanzi and Rolland, Chloe and Gustafson, Laura and Xiao, Tete and Whitehead, Spencer and Berg, Alexander C and Lo, Wan-Yen and others},
  booktitle={Proceedings of the IEEE/CVF International Conference on Computer Vision},
  pages={4015--4026},
  year={2023}
}

@inproceedings{he2023weakly,
  title={Weakly-supervised camouflaged object detection with scribble annotations},
  author={He, Ruozhen and Dong, Qihua and Lin, Jiaying and Lau, Rynson WH},
  booktitle={Proceedings of the AAAI Conference on Artificial Intelligence},
  volume={37},
  pages={781--789},
  year={2023}
}

@article{he2024weakly,
  title={Weakly-supervised concealed object segmentation with sam-based pseudo labeling and multi-scale feature grouping},
  author={He, Chunming and Li, Kai and Zhang, Yachao and Xu, Guoxia and Tang, Longxiang and Zhang, Yulun and Guo, Zhenhua and Li, Xiu},
  journal={Advances in Neural Information Processing Systems},
  volume={36},
  year={2023}
}

@article{mei2023distraction,
  title={Distraction-aware camouflaged object segmentation},
  author={Mei, Haiyang and Yang, Xin and Zhou, Yunduo and Ji, Ge-Peng and Wei, Xiaopeng and Fan, DP},
  journal={SCIENTIA SINICA Informationis (SSI)},
  volume={3},
  pages={7},
  year={2023}
}

@article{xu2021boundary,
  title={Boundary guidance network for camouflage object detection},
  author={Xu, Xiuqi and Zhu, Mingyu and Yu, Jinhao and Chen, Shuhan and Hu, Xuelong and Yang, Yuequan},
  journal={Image and Vision Computing},
  volume={114},
  pages={104283},
  year={2021},
  publisher={Elsevier}
}

@article{ji2023deep,
  title={Deep gradient learning for efficient camouflaged object detection},
  author={Ji, Ge-Peng and Fan, Deng-Ping and Chou, Yu-Cheng and Dai, Dengxin and Liniger, Alexander and Van Gool, Luc},
  journal={Machine Intelligence Research},
  volume={20},
  number={1},
  pages={92--108},
  year={2023},
  publisher={Springer}
}

@article{liu2023mscaf,
  title={MSCAF-net: A general framework for camouflaged object detection via learning multi-scale context-aware features},
  author={Liu, Yu and Li, Haihang and Cheng, Juan and Chen, Xun},
  journal={IEEE Transactions on Circuits and Systems for Video Technology},
  volume={33},
  number={9},
  pages={4934--4947},
  year={2023},
  publisher={IEEE}
}

@article{yue2023dual,
  title={Dual-constraint coarse-to-fine network for camouflaged object detection},
  author={Yue, Guanghui and Xiao, Houlu and Xie, Hai and Zhou, Tianwei and Zhou, Wei and Yan, Weiqing and Zhao, Baoquan and Wang, Tianfu and Jiang, Qiuping},
  journal={IEEE Transactions on Circuits and Systems for Video Technology},
  year={2023},
  publisher={IEEE}
}

@inproceedings{zhang2020weakly,
  title={Weakly-supervised salient object detection via scribble annotations},
  author={Zhang, Jing and Yu, Xin and Li, Aixuan and Song, Peipei and Liu, Bowen and Dai, Yuchao},
  booktitle={Proceedings of the IEEE/CVF conference on computer vision and pattern recognition},
  pages={12546--12555},
  year={2020}
}

@inproceedings{yu2021structure,
  title={Structure-consistent weakly supervised salient object detection with local saliency coherence},
  author={Yu, Siyue and Zhang, Bingfeng and Xiao, Jimin and Lim, Eng Gee},
  booktitle={Proceedings of the AAAI conference on artificial intelligence},
  volume={35},
  pages={3234--3242},
  year={2021}
}

@inproceedings{liang2022tree,
  title={Tree energy loss: Towards sparsely annotated semantic segmentation},
  author={Liang, Zhiyuan and Wang, Tiancai and Zhang, Xiangyu and Sun, Jian and Shen, Jianbing},
  booktitle={Proceedings of the IEEE/CVF conference on computer vision and pattern recognition},
  pages={16907--16916},
  year={2022}
}

@article{liu2021weakly,
  title={Weakly-supervised salient object detection with saliency bounding boxes},
  author={Liu, Yuxuan and Wang, Pengjie and Cao, Ying and Liang, Zijian and Lau, Rynson WH},
  journal={IEEE Transactions on Image Processing},
  volume={30},
  pages={4423--4435},
  year={2021},
  publisher={IEEE}
}

@article{wang2023weakly,
  title={Weakly supervised salient object detection algorithm based on bounding box annotation},
  author={WANG, Qiang and HUANG, Xiaoming and TONG, Qiang and LIU, Xiulei},
  journal={Journal of Computer Applications},
  volume={43},
  number={6},
  pages={1910},
  year={2023}
}

@article{ji2304sam,
  title={SAM struggles in concealed scenes—empirical study on “Segment Anything”},
  author={Ji, Ge-Peng and Fan, Deng-Ping and Xu, Peng and Zhou, Bowen and Cheng, Ming-Ming and Van Gool, Luc},
  journal={Science China Information Sciences},
  volume={66},
  number={12},
  pages={226101},
  year={2023},
  publisher={Springer}
}

@article{le2019anabranch,
  title={Anabranch network for camouflaged object segmentation},
  author={Le, Trung-Nghia and Nguyen, Tam V and Nie, Zhongliang and Tran, Minh-Triet and Sugimoto, Akihiro},
  journal={Computer vision and image understanding},
  volume={184},
  pages={45--56},
  year={2019},
  publisher={Elsevier}
}

@inproceedings{wang2021pyramid,
  title={Pyramid vision transformer: A versatile backbone for dense prediction without convolutions},
  author={Wang, Wenhai and Xie, Enze and Li, Xiang and Fan, Deng-Ping and Song, Kaitao and Liang, Ding and Lu, Tong and Luo, Ping and Shao, Ling},
  booktitle={Proceedings of the IEEE/CVF international conference on computer vision},
  pages={568--578},
  year={2021}
}

@article{wang2022pvt,
  title={Pvt v2: Improved baselines with pyramid vision transformer},
  author={Wang, Wenhai and Xie, Enze and Li, Xiang and Fan, Deng-Ping and Song, Kaitao and Liang, Ding and Lu, Tong and Luo, Ping and Shao, Ling},
  journal={Computational Visual Media},
  volume={8},
  number={3},
  pages={415--424},
  year={2022},
  publisher={Springer}
}

@inproceedings{woo2018cbam,
  title={Cbam: Convolutional block attention module},
  author={Woo, Sanghyun and Park, Jongchan and Lee, Joon-Young and Kweon, In So},
  booktitle={Proceedings of the European conference on computer vision (ECCV)},
  pages={3--19},
  year={2018}
}

@inproceedings{wei2020f3net,
  title={F$^3$Net: fusion, feedback and focus for salient object detection},
  author={Wei, Jun and Wang, Shuhui and Huang, Qingming},
  booktitle={Proceedings of the AAAI conference on artificial intelligence},
  volume={34},
  pages={12321--12328},
  year={2020}
}

@inproceedings{lv2021simultaneously,
  title={Simultaneously localize, segment and rank the camouflaged objects},
  author={Lv, Yunqiu and Zhang, Jing and Dai, Yuchao and Li, Aixuan and Liu, Bowen and Barnes, Nick and Fan, Deng-Ping},
  booktitle={Proceedings of the IEEE/CVF conference on computer vision and pattern recognition},
  pages={11591--11601},
  year={2021}
}

@inproceedings{wang2017learning,
  title={Learning to detect salient objects with image-level supervision},
  author={Wang, Lijun and Lu, Huchuan and Wang, Yifan and Feng, Mengyang and Wang, Dong and Yin, Baocai and Ruan, Xiang},
  booktitle={Proceedings of the IEEE conference on computer vision and pattern recognition},
  pages={136--145},
  year={2017}
}

@inproceedings{yan2013hierarchical,
  title={Hierarchical saliency detection},
  author={Yan, Qiong and Xu, Li and Shi, Jianping and Jia, Jiaya},
  booktitle={Proceedings of the IEEE conference on computer vision and pattern recognition},
  pages={1155--1162},
  year={2013}
}

@inproceedings{yang2013saliency,
  title={Saliency detection via graph-based manifold ranking},
  author={Yang, Chuan and Zhang, Lihe and Lu, Huchuan and Ruan, Xiang and Yang, Ming-Hsuan},
  booktitle={Proceedings of the IEEE conference on computer vision and pattern recognition},
  pages={3166--3173},
  year={2013}
}

@article{li2016visual,
  title={Visual saliency detection based on multiscale deep CNN features},
  author={Li, Guanbin and Yu, Yizhou},
  journal={IEEE transactions on image processing},
  volume={25},
  number={11},
  pages={5012--5024},
  year={2016},
  publisher={IEEE}
}

@inproceedings{yang2021uncertainty,
  title={Uncertainty-guided transformer reasoning for camouflaged object detection},
  author={Yang, Fan and Zhai, Qiang and Li, Xin and Huang, Rui and Luo, Ao and Cheng, Hong and Fan, Deng-Ping},
  booktitle={Proceedings of the IEEE/CVF international conference on computer vision},
  pages={4146--4155},
  year={2021}
}

@article{fan2021concealed,
  title={Concealed object detection},
  author={Fan, Deng-Ping and Ji, Ge-Peng and Cheng, Ming-Ming and Shao, Ling},
  journal={IEEE transactions on pattern analysis and machine intelligence},
  volume={44},
  number={10},
  pages={6024--6042},
  year={2021},
  publisher={IEEE}
}

@article{yan2023camouflaged,
  title={Camouflaged object segmentation based on matching--recognition--refinement network},
  author={Yan, Xinyu and Sun, Meijun and Han, Yahong and Wang, Zheng},
  journal={IEEE Transactions on Neural Networks and Learning Systems},
  year={2023},
  publisher={IEEE}
}

@inproceedings{khan2024camofocus,
  title={CamoFocus: Enhancing Camouflage Object Detection With Split-Feature Focal Modulation and Context Refinement},
  author={Khan, Abbas and Khan, Mustaqeem and Gueaieb, Wail and El Saddik, Abdulmotaleb and De Masi, Giulia and Karray, Fakhri},
  booktitle={Proceedings of the IEEE/CVF Winter Conference on Applications of Computer Vision},
  pages={1434--1443},
  year={2024}
}

@inproceedings{he2023strategic,
  title={Strategic preys make acute predators: Enhancing camouflaged object detectors by generating camouflaged objects},
  author={He, Chunming and Li, Kai and Zhang, Yachao and Zhang, Yulun and Guo, Zhenhua and Li, Xiu and Danelljan, Martin and Yu, Fisher},
  booktitle={International Conference on Learning Representations},
  year={2024}
}

@article{wang2024ipnet,
  title={IPNet: Polarization-based Camouflaged Object Detection via dual-flow network},
  author={Wang, Xin and Ding, Jiajia and Zhang, Zhao and Xu, Junfeng and Gao, Jun},
  journal={Engineering Applications of Artificial Intelligence},
  volume={127},
  pages={107303},
  year={2024},
  publisher={Elsevier}
}

@inproceedings{liu2022boosting,
  title={Boosting camouflaged object detection with dual-task interactive transformer},
  author={Liu, Zhengyi and Zhang, Zhili and Tan, Yacheng and Wu, Wei},
  booktitle={2022 26th International Conference on Pattern Recognition (ICPR)},
  pages={140--146},
  year={2022},
  organization={IEEE}
}

@inproceedings{hu2023high,
  title={High-resolution iterative feedback network for camouflaged object detection},
  author={Hu, Xiaobin and Wang, Shuo and Qin, Xuebin and Dai, Hang and Ren, Wenqi and Luo, Donghao and Tai, Ying and Shao, Ling},
  booktitle={Proceedings of the AAAI Conference on Artificial Intelligence},
  volume={37},
  pages={881--889},
  year={2023}
}

@article{yang2024spatial,
  title={Spatial Coherence Loss for Salient and Camouflaged Object Detection and Beyond},
  author={Yang, Ziyun and Choy, Kevin and Farsiu, Sina},
  journal={arXiv preprint arXiv:2402.18698},
  year={2024}
}

@inproceedings{shi2022ssformer,
  title={Ssformer: A lightweight transformer for semantic segmentation},
  author={Shi, Wentao and Xu, Jing and Gao, Pan},
  booktitle={2022 IEEE 24th International Workshop on Multimedia Signal Processing (MMSP)},
  pages={1--5},
  year={2022},
  organization={IEEE}
}

@inproceedings{jha2020kvasir,
  title={Kvasir-seg: A segmented polyp dataset},
  author={Jha, Debesh and Smedsrud, Pia H and Riegler, Michael A and Halvorsen, P{\aa}l and De Lange, Thomas and Johansen, Dag and Johansen, H{\aa}vard D},
  booktitle={MultiMedia modeling: 26th international conference, MMM 2020, Daejeon, South Korea, January 5--8, 2020, proceedings, part II 26},
  pages={451--462},
  year={2020},
  organization={Springer}
}

@article{bernal2015wm,
  title={WM-DOVA maps for accurate polyp highlighting in colonoscopy: Validation vs. saliency maps from physicians},
  author={Bernal, Jorge and S{\'a}nchez, F Javier and Fern{\'a}ndez-Esparrach, Gloria and Gil, Debora and Rodr{\'\i}guez, Cristina and Vilari{\~n}o, Fernando},
  journal={Computerized medical imaging and graphics},
  volume={43},
  pages={99--111},
  year={2015},
  publisher={Elsevier}
}

@article{tajbakhsh2015automated,
  title={Automated polyp detection in colonoscopy videos using shape and context information},
  author={Tajbakhsh, Nima and Gurudu, Suryakanth R and Liang, Jianming},
  journal={IEEE transactions on medical imaging},
  volume={35},
  number={2},
  pages={630--644},
  year={2015},
  publisher={IEEE}
}

@article{tang2023can,
  title={Can sam segment anything? when sam meets camouflaged object detection},
  author={Tang, Lv and Xiao, Haoke and Li, Bo},
  journal={arXiv preprint arXiv:2304.04709},
  year={2023}
}

@article{gao2019res2net,
  title={Res2net: A new multi-scale backbone architecture},
  author={Gao, Shang-Hua and Cheng, Ming-Ming and Zhao, Kai and Zhang, Xin-Yu and Yang, Ming-Hsuan and Torr, Philip},
  journal={IEEE transactions on pattern analysis and machine intelligence},
  volume={43},
  number={2},
  pages={652--662},
  year={2019},
  publisher={IEEE}
}

@inproceedings{zhu2022can,
  title={I can find you! boundary-guided separated attention network for camouflaged object detection},
  author={Zhu, Hongwei and Li, Peng and Xie, Haoran and Yan, Xuefeng and Liang, Dong and Chen, Dapeng and Wei, Mingqiang and Qin, Jing},
  booktitle={Proceedings of the AAAI conference on artificial intelligence},
  volume={36},
  pages={3608--3616},
  year={2022}
}

@inproceedings{rahman2023medical,
  title={Medical image segmentation via cascaded attention decoding},
  author={Rahman, Md Mostafijur and Marculescu, Radu},
  booktitle={Proceedings of the IEEE/CVF Winter Conference on Applications of Computer Vision},
  pages={6222--6231},
  year={2023}
}

@article{yin2024camoformer,
  title={Camoformer: Masked separable attention for camouflaged object detection},
  author={Yin, Bowen and Zhang, Xuying and Fan, Deng-Ping and Jiao, Shaohui and Cheng, Ming-Ming and Van Gool, Luc and Hou, Qibin},
  journal={IEEE Transactions on Pattern Analysis and Machine Intelligence},
  year={2024},
  publisher={IEEE}
}

@article{chen2024sam,
  title={SAM-COD: SAM-guided Unified Framework for Weakly-Supervised Camouflaged Object Detection},
  author={Chen, Huafeng and Wei, Pengxu and Guo, Guangqian and Gao, Shan},
  journal={European Conference on Computer Vision},
  year={2024},
  publisher={IEEE}
}

@misc{ji2024segment,
  title={Segment anything is not always perfect: An investigation of sam on different real-world applications},
  author={Ji, Wei and Li, Jingjing and Bi, Qi and Liu, Tingwei and Li, Wenbo and Cheng, Li},
  year={2024},
  publisher={Springer}
}

@article{wang2021salient,
  title={Salient object detection in the deep learning era: An in-depth survey},
  author={Wang, Wenguan and Lai, Qiuxia and Fu, Huazhu and Shen, Jianbing and Ling, Haibin and Yang, Ruigang},
  journal={IEEE Transactions on Pattern Analysis and Machine Intelligence},
  volume={44},
  number={6},
  pages={3239--3259},
  year={2021},
  publisher={IEEE}
}

@article{xu2023guided,
  title={Guided multi-scale refinement network for camouflaged object detection},
  author={Xu, Xiuqi and Chen, Shuhan and Lv, Xiao and Wang, Jian and Hu, Xuelong},
  journal={Multimedia Tools and Applications},
  volume={82},
  number={4},
  pages={5785--5801},
  year={2023},
  publisher={Springer}
}

@inproceedings{li2018csrnet,
  title={Csrnet: Dilated convolutional neural networks for understanding the highly congested scenes},
  author={Li, Yuhong and Zhang, Xiaofan and Chen, Deming},
  booktitle={Proceedings of the IEEE conference on computer vision and pattern recognition},
  pages={1091--1100},
  year={2018}
}

@inproceedings{gao2022weakly,
  title={Weakly supervised video salient object detection via point supervision},
  author={Gao, Shuyong and Xing, Haozhe and Zhang, Wei and Wang, Yan and Guo, Qianyu and Zhang, Wenqiang},
  booktitle={Proceedings of the 30th ACM International Conference on Multimedia},
  pages={3656--3665},
  year={2022}
}

@inproceedings{zhang2023input,
  title={Input augmentation with sam: Boosting medical image segmentation with segmentation foundation model},
  author={Zhang, Yizhe and Zhou, Tao and Wang, Shuo and Liang, Peixian and Zhang, Yejia and Chen, Danny Z},
  booktitle={International Conference on Medical Image Computing and Computer-Assisted Intervention},
  pages={129--139},
  year={2023},
  organization={Springer}
}

@article{chen2024ma,
  title={Ma-sam: Modality-agnostic sam adaptation for 3d medical image segmentation},
  author={Chen, Cheng and Miao, Juzheng and Wu, Dufan and Zhong, Aoxiao and Yan, Zhiling and Kim, Sekeun and Hu, Jiang and Liu, Zhengliang and Sun, Lichao and Li, Xiang and others},
  journal={Medical Image Analysis},
  pages={103310},
  year={2024},
  publisher={Elsevier}
}

@article{zhang2021bilateral,
  title={Bilateral attention network for RGB-D salient object detection},
  author={Zhang, Zhao and Lin, Zheng and Xu, Jun and Jin, Wen-Da and Lu, Shao-Ping and Fan, Deng-Ping},
  journal={IEEE transactions on image processing},
  volume={30},
  pages={1949--1961},
  year={2021},
  publisher={IEEE}
}

@inproceedings{zhang2025learning,
  title={Learning Camouflaged Object Detection from Noisy Pseudo Label},
  author={Zhang, Jin and Zhang, Ruiheng and Shi, Yanjiao and Cao, Zhe and Liu, Nian and Khan, Fahad Shahbaz},
  booktitle={European Conference on Computer Vision},
  pages={158--174},
  year={2025},
  organization={Springer}
}

@article{chen2017deeplab,
  title={Deeplab: Semantic image segmentation with deep convolutional nets, atrous convolution, and fully connected crfs},
  author={Chen, Liang-Chieh and Papandreou, George and Kokkinos, Iasonas and Murphy, Kevin and Yuille, Alan L},
  journal={IEEE transactions on pattern analysis and machine intelligence},
  volume={40},
  number={4},
  pages={834--848},
  year={2017},
  publisher={IEEE}
}

@inproceedings{zhao2024focusdiffuser,
  title={Focusdiffuser: Perceiving local disparities for camouflaged object detection},
  author={Zhao, Jianwei and Li, Xin and Yang, Fan and Zhai, Qiang and Luo, Ao and Jiao, Zicheng and Cheng, Hong},
  booktitle={European Conference on Computer Vision},
  pages={181--198},
  year={2024},
  organization={Springer}
}

@article{liu2025ssfam,
  title={SSFam: Scribble Supervised Salient Object Detection Family},
  author={Liu, Zhengyi and Deng, Sheng and Wang, Xinrui and Wang, Linbo and Fang, Xianyong and Tang, Bin},
  journal={IEEE Transactions on Multimedia},
  year={2025},
  publisher={IEEE}
}

@incollection{chen2023diffusion,
  title={Diffusion model for camouflaged object detection},
  author={Chen, Zhennan and Gao, Rongrong and Xiang, Tian-Zhu and Lin, Fan},
  booktitle={ECAI 2023},
  pages={445--452},
  year={2023},
  publisher={IOS Press}
}

@inproceedings{yu2024exploring,
  title={Exploring Deeper! Segment Anything Model with Depth Perception for Camouflaged Object Detection},
  author={Yu, Zhenni and Zhang, Xiaoqin and Zhao, Li and Bin, Yi and Xiao, Guobao},
  booktitle={Proceedings of the 32nd ACM International Conference on Multimedia},
  pages={4322--4330},
  year={2024}
}

@article{liu2024search,
  title={Search and recovery network for camouflaged object detection},
  author={Liu, Guangrui and Wu, Wei},
  journal={Image and Vision Computing},
  volume={151},
  pages={105247},
  year={2024},
  publisher={Elsevier}
}

@article{ye2024reverse,
  title={Reverse cross-refinement network for camouflaged object detection},
  author={Ye, Qian and Zhou, Yaqin and Huo, Guanying and Liu, Yan and Zhou, Yan and Li, Qingwu},
  journal={Image and Vision Computing},
  volume={150},
  pages={105218},
  year={2024},
  publisher={Elsevier}
}

@article{zheng2024bilateral,
  title={Bilateral Reference for High-Resolution Dichotomous Image Segmentation},
  author={Zheng, Peng and Gao, Dehong and Fan, Deng-Ping and Liu, Li and Laaksonen, Jorma and Ouyang, Wanli and Sebe, Nicu},
  journal={CAAI Artificial Intelligence Research},
  volume={3},
  year={2024},
}

@article{pang2024zoomnext,
  title={Zoomnext: A unified collaborative pyramid network for camouflaged object detection},
  author={Pang, Youwei and Zhao, Xiaoqi and Xiang, Tian-Zhu and Zhang, Lihe and Lu, Huchuan},
  journal={IEEE transactions on pattern analysis and machine intelligence},
  year={2024},
  publisher={IEEE}
}

@inproceedings{zhang2018image,
  title={Image super-resolution using very deep residual channel attention networks},
  author={Zhang, Yulun and Li, Kunpeng and Li, Kai and Wang, Lichen and Zhong, Bineng and Fu, Yun},
  booktitle={Proceedings of the European conference on computer vision (ECCV)},
  pages={286--301},
  year={2018}
}

@inproceedings{liu2022adaptive,
  title={Adaptive early-learning correction for segmentation from noisy annotations},
  author={Liu, Sheng and Liu, Kangning and Zhu, Weicheng and Shen, Yiqiu and Fernandez-Granda, Carlos},
  booktitle={Proceedings of the IEEE/CVF Conference on Computer Vision and Pattern Recognition},
  pages={2606--2616},
  year={2022}
}



\end{document}